\documentclass[sigconf]{acmart}

\usepackage{times}
\usepackage{soul}
\usepackage{url}
\usepackage[utf8]{inputenc}
\usepackage{caption}
\usepackage{graphicx}
\usepackage{amsmath}
\usepackage{amsthm}
\usepackage{booktabs}
\usepackage{algorithm}
\usepackage{algorithmic}
\usepackage{float}
\usepackage{hyperref}
\usepackage[switch]{lineno}

\newtheorem{definition}{Definition}
\newtheorem{corollary}{Corollary}
\newtheorem{theorem}{Theorem}

\title{HoGA: Higher-Order Graph Attention via\\ Diversity-Aware k-Hop Sampling}

\author{Thomas Bailie}
\affiliation{%
  \institution{School of Computer Science\\The University of Auckland}
  \city{Auckland}
  \country{New Zealand}
}
\email{thomas.bailie@auckland.ac.nz}

\author{Yun Sing Koh}
\affiliation{%
  \institution{School of Computer Science\\The University of Auckland}
  \city{Auckland}
  \country{New Zealand}
}
\email{y.koh@auckland.ac.nz}

\author{Surya Karthik Mukkavilli}
\affiliation{%
  \institution{Mercuria Energy Group}
  \city{Geneva}
  \country{Switzerland}
}
\affiliation{%
  \institution{Graduate School of Green Growth\\Korea Advanced Institute of Science and Technology}
  \city{Daejeon}
  \country{South Korea}
}
\email{drkarthik@kaist.ac.kr}

\usepackage{amsmath}
\newcommand{\best}[1]{\mathbf{#1}}

\AtBeginDocument{%
  }

\copyrightyear{2026}
\acmYear{2026}
\acmDOI{10.1145/3773966.3777960}
\acmConference[WSDM ’26]{The 19th ACM International Conference on Web Search and Data Mining}{February 22--26, 2026}{Boise, Idaho, USA}

\usepackage{amsmath}
\usepackage{amsfonts}
\usepackage{booktabs}
\usepackage{array}
\usepackage{float}
\usepackage{xcolor} 



\begin{document}

\begin{abstract}
    Graphs model latent variable relationships in many real-world systems, and Message Passing Neural Networks (MPNNs) are widely used to learn such structures for downstream tasks. While edge-based MPNNs effectively capture local interactions, their expressive power is theoretically bounded, limiting the discovery of higher-order relationships. We introduce the Higher-Order Graph Attention (HoGA) module, which constructs a $k$-order attention matrix by sampling subgraphs to maximize diversity among feature vectors. Unlike existing higher-order attention methods that greedily resample similar $k$-order relationships, HoGA targets diverse modalities in higher-order topology, reducing redundancy and expanding the range of captured substructures. Applied to two single-hop attention models, HoGA achieves at least a 5\% accuracy gain on all benchmark node classification datasets and outperforms recent baselines on six of eight datasets. Code is available at \url{https://github.com/TB862/Higher_Order}.
\end{abstract}

\maketitle
\title{HoGA: Higher-Order Graph Attention via Diversity-Aware k-Hop Sampling}
%

\section{Introduction}

Message Passing Neural Networks (MPNNs) capture latent relationships relevant to downstream tasks \cite{scarselli2009graph} and have been successfully applied in diverse domains, including molecular chemistry \cite{gilmer2017mpnn}, transport planning \cite{jiang2022graph}, social networks \cite{fan2019graph}, drug discovery \cite{xiong2021graph}, and climate modeling \cite{lam2023graphcast}. However, strictly edge-based MPNNs neglect global relationships embedded in the graph topology. This locality bias introduces bottlenecks \cite{topping2021understanding,di2023over} that result in the loss of higher-order structural signals \cite{ai2024a2gcn} and impose theoretical limits on the patterns they can detect \cite{xu2018powerful}. As a result, single-hop models fail to identify intricate topological substructures \cite{zhang2023rethinking}. For example, in social networks, association groups with intricate topology often remain undetected by edge-based MPNNs \cite{xiong2024graph}. \citet{guyan2025pegnn} and \citet{liu2024segcn} attribute the difficulty of detecting coordinated bot networks to this omission of higher-order relational information.

\begin{figure}[t]
    \centering
    \vspace{20pt}
    \includegraphics[width=\columnwidth]{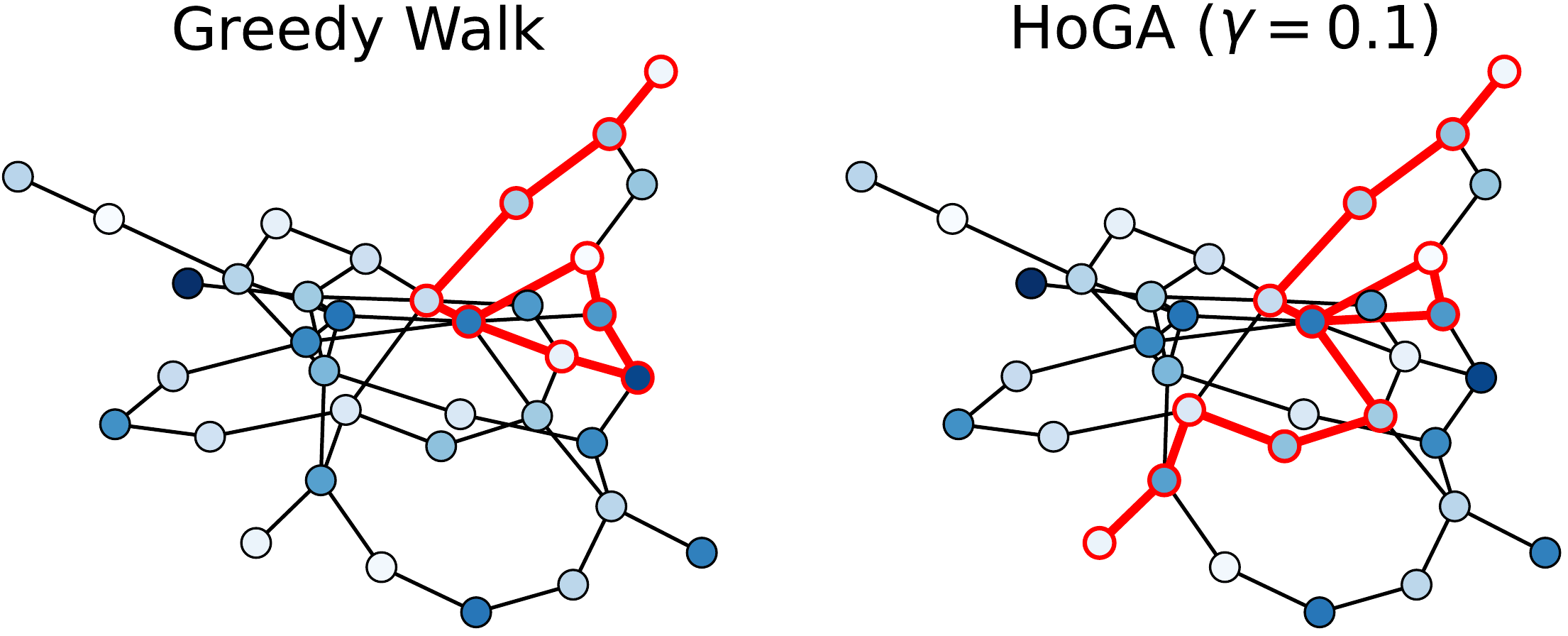}
    \caption{A $40$ node undirected random graph, initialized via the Erdos–Renyi algorithm. The probability of edge existence is set to $8\%$, while the random seed is $41$. Coloring represents the magnitude of feature-vector values between $0$ and $1$. Shown in bold red edges are $15$ steps of a greedy walk and the heuristically guided walk of HoGA. These walks aim to maximize diversity. Greedy is prone to getting stuck in cycles, while HoGA is able to escape them, given its diversity heuristic.}
    \label{fig:front_fig}
\end{figure}

To address the locality limitation, recent high-order attention methods assign weights to $k$-simplexes  \cite{huang2024higher} and $k$-neighborhoods \cite{zhang2024hongat}, or sample meta-paths originating from each node \cite{yang2021spagan,yang2016revisiting}. However, though these approaches excel at sampling within their problem contexts, they invariably rely on greedy sampling from a distribution of possible higher-order relationships. Such strategies often redraw previously selected samples, limiting the effective exploration of their intended search space. Consequently, the model’s ability to uncover latent topological structures is diminished \cite{zhang2023rethinking}. This limitation is exacerbated by the fact that the state space of possible $k$-hop relationships grows exponentially with $k$, making exhaustive sampling infeasible. Under a fixed sampling budget, greediness tends to bias selection toward a narrow subset of $k$-hop feature-space modalities, further under-representing higher-order concepts.

To improve the discovery of latent topological structures, we propose the Higher-Order Graph Attention (HoGA) module, designed to capture global $k$-hop relationships even under stringent sampling budgets. HoGA assigns weights to higher-order connections by sampling fixed substructures, enriching the representation of global dependencies within the model’s latent reasoning. Specifically, HoGA constructs subgraphs of the $k$-hop neighborhood through a heuristically guided walk (Figure~\ref{fig:front_fig} provides example walks). HoGA's walk balances global and local diversity metrics, enabling high-fidelity approximation of the otherwise intractable $k$-hop state space via iterative sampling. Unlike conventional approaches, HoGA avoids revisiting previously explored $k$-hop features, thereby collecting distinct modalities in the underlying graph topology and enhancing their diversity. This diversity-driven heuristic increases coverage of the $k$-hop feature space, improving the model’s ability to detect association groups and other intricate topological substructures.


We empirically evaluate the integration of the Higher-Order Graph Attention (HoGA) module into two distinct attention-based models, GAT \cite{velickovic2018graph} and GRAND \cite{chamberlain2021grand}, across both homophilic and heterophilic benchmark node classification datasets \cite{yang2016revisiting}. Our results show that HoGA-enhanced models (HoGA-GAT and HoGA-GRAND) achieve substantial accuracy gains over their respective base models. Relative to other baselines, they achieve statistically significant gains on six out of eight datasets evaluated on.

Our contributions are summarized as follows.
(1) We propose HoGA, a graph attention module that samples $k$-hop variable relationships via heuristically guided walks. By promoting sampling diversity, HoGA enables high-fidelity estimation of the $k$-hop feature space.
(2) We show in our theoretical analysis that HoGA reduces redundant sampling of higher-order graph substructures, while our empirical results demonstrate its ability to mitigate the performance degradation commonly observed in deeper networks.
(3) HoGA extends single-hop attention MPNNs to the $k$-hop setting. We show through rigorous experimentation that applying HoGA to existing single-hop attention models, namely GAT \cite{velickovic2018graph} and GRAND \cite{chamberlain2021grand}, yields significant accuracy improvements over both homophilic and heterophilic node classification benchmarks datasets.

\section{Related Work}

\begin{figure*}[t]
    \centering
    \includegraphics[width=2\columnwidth]{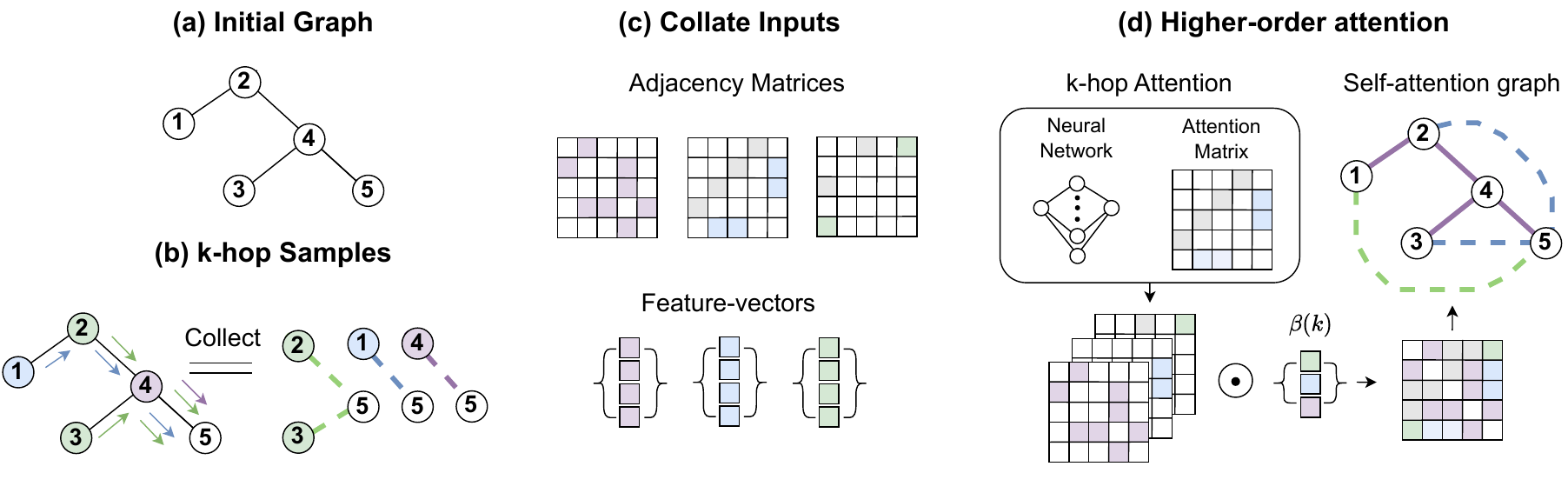}
    \caption{The Higher-order Graphical Attention (HoGA) module. {\bf (a)} an input graph of arbitrary topology. {\bf (b)} HoGA samples the $k$-hop neighborhood up to a maximum value of $K$ via a heuristic walk. {\bf (c)} The sampling results create an adjacency matrix describing connections via a shortest path of length $k$. {\bf (d)} Higher-order aggregation combines nodal information of variable distance, thus recreating the initial graph with self-attention edge weights.}
    \label{fig:walk_proc}
\end{figure*}

Message-passing regimes that emphasize strict locality suffer from several well-known issues, including oversmoothing, oversquashing, and provably limited expressive power~\cite{morris2019weisfeiler}.

Since the update function in MPNNs is generally not injective, the expressive power of first-order message passing is inherently limited. The set of non-isomorphic graphs that any $1$-hop aggregation scheme can uniquely color is a strict subset of those distinguishable by the $1$-WL test \cite{morris2019weisfeiler}. Enhancing MPNN expressivity beyond the $1$-WL test has therefore become a critical research area. \citet{zhang2023rethinking} show that MPNNs aggregate feature vectors across all $k$-hops and exhibit expressivity between $1$- and $3$-WL.

Due to the finite capacity of hidden representations, long-range information is gradually lost through message passing. Analytically, oversquashing can be quantified in terms of information bottleneck severity \cite{di2023over,topping2021understanding}. To mitigate this, graph rewiring techniques either add \cite{gutteridge2023drew,arnaiz2022diffwire} or remove \cite{karhadkar2022fosr} edges to improve information flow. However, such rewiring does not explicitly preserve the original topology and may create superficial connections while removing genuine ones. In contrast, $k$-hop aggregation maintains topology while reducing the effective commute time between any two nodes by a factor of $k$, thereby theoretically alleviating oversquashing \cite{topping2021understanding}.

Overlapping receptive fields cause node feature vectors to converge toward constant values, corresponding to the Dirichlet energy approaching zero \cite{rusch2023survey}. In contrast, feature diversity between nodes tends to increase with hop distance within the $k$-hop neighborhood \cite{ai2024a2gcn}. \citet{wang2020multi} shows that re-weighting the adjacency matrix based on multi-hop connectivity mitigates oversmoothing and improves performance.

Single-hop attention schemes are well studied \cite{kipf2016semi,xu2018powerful,gilmer2017mpnn,chamberlain2021grand}. Notably, \citet{velickovic2018graph} introduced self-attention on graph edges. Recent work has extended attention to paths of arbitrary length, treating edges as paths of length one \cite{zhang2024hongat,huang2024higher}. However, the non-polynomial growth of the $k$-hop state space remains a major challenge when aggregating higher-order feature vectors. Several approaches aim to leverage long-range relationships while maintaining tractability. For example, Abboud et al.~\cite{abboud2022shortest} map the entire $k$-hop neighborhood into a single feature vector via an injective aggregation function. Yet, due to the inherent diversity of $k$-hop features, such aggregation risks oversmoothing distinct modalities.
Walk-based approaches \cite{yeh2023random,kong2022geodesic,li2020distance,michel2023path} sample the $k$-hop neighborhood by constructing length-$k$ paths. For instance, Michel et al.\cite{michel2023path} aggregate samples from the set of shortest and simple paths between node pairs into a single feature vector, while Kong et al.\cite{kong2022geodesic} pool features from nodes along shortest paths by concatenating degree embeddings. These methods offer a flexible multi-hop paradigm, reducing \textit{a priori} topological constraints while avoiding exponential state-space growth.

Despite these advances, little attention has been given to methods that directly operate on the complete set of $k$-hop neighbors—despite their theoretical advantages for uncovering richer topological substructures \cite{zhang2023rethinking}. To address this gap, we propose a graph attention module that iteratively samples from the global $k$-hop feature space. Our experiments demonstrate significant performance improvements over state-of-the-art higher-order attention models.

\section{Preliminaries}

A key component of GNNs is their ability to directly learn fixed ground truth topology over a non-Euclidean domain. Here, we formulate the notion of local message-passing operators on graphs and the node classification task. 

\noindent{\bf Message Passing Graph Neural Networks.}
Let $G = (V, E)$ be a fully connected undirected graph with vertices $V$ and edges $E$. Intuitively, $i, j \in V$ are variables from a state space over a real-world domain, {\it e.g.}, individuals or papers within social or citation networks. The topology described within $E$ defines the inherent relationships between these variables: if $(i, j) \in E$, $i$, and $j$ are directly related. Alternatively, should a path $\mathcal{P} = (i_1, \dots, i_k)$ of length $k$ exist between $i_1=i$ and $i_k=j$, we denote $i$ and $j$ as having a relation by casualty of the nodes in the path $i_1, \dots, i_k \in V$. Furthermore, owing to the sequential dependence on $k$ variables, the relationship between $\mathbf{x}_{i_k}$ and $\mathbf{x}_{i_1}$ is of order $k$. Correspondingly, each variable $i \in V$ is associated with a {\it feature-vector} $\mathbf{x}_i(t)$ at depth $t$. 

MPNNs are information aggregation schemes that concurrently aggregate nodal features across the whole graph. Notably, initial schemes, such as those used in \citet{kipf2016semi,xu2018powerful,hamilton2017inductive} and \citet{velickovic2018graph}, focused on aggregation of node $i$'s direct neighbors at a shortest path distance of one from $i$. The general form of any single-hop MPNN layer is given as:
\begin{equation}
    \mathbf{x}_i(t+1) = \Psi_t\left(\mathbf{x}_i(t), \phi_t\left(\{\!\{ \mathbf{x}_j(t) \mid j \in \mathcal{N}_1(i) \}\!\}\right)\right),
    \label{eq:1agg}
\end{equation}
where we have denoted the multi-set as $\{\!\{\}\!\}$ and the nodes with shortest path length $k$ from $i$ as $\mathcal{N}_k(i)$. Here, $\phi_t$ is an injective function aggregating close-proximity feature vectors, thus allowing for a tractable update function $\Psi_t$ to create a new feature representation. An example of $\phi_t$ is the summation operator $\sum_i \mathbf{x}_i(t)$. 

\noindent{\bf Node Classification.}
Each $i \in V$ has an associated ground truth label $y_i$ for a node classification task. In the semi-supervised setting, the MPNN classifier predicts the classes of a subset $\mathcal{S} \subset V$, which corresponds to the training, testing, or validation sets. Note that the union of all three does not necessarily contain all $i \in V$, and is often a small subset. To train our network,  we use the Cross Entropy Loss function, which for output logits $\hat{y}_i$ is given as: 
\begin{equation}
    \mathcal{L}_{}(y, \hat{y}) = - \sum_{l=1}^{\left\vert S \right\vert} y_l \log(\hat{y}_l).
\end{equation}

\noindent{\bf Single-hop Graphical Attention.}
In the single-hop setting, attention weights $\alpha_{i,j}$ were strictly considered to be along edges $(i, j) \in E$ \cite{velickovic2018graph}. The corresponding attention matrix $\mathbf{A}(\mathbf{x}(t))$ is therefore only non-zero on edges of the graph:
\begin{equation}
    \label{eq:attn}
    \mathbf{A}(\mathbf{x}(t))_{i,j} = 
    \begin{cases}
        \displaystyle \alpha_{i, j} & \text{if } (i, j) \in E, \\
        \displaystyle 0 & \text{otherwise}.
    \end{cases}
\end{equation}
Here, $\alpha_{i,j}$ is a normalised attention coefficient calculated by learning the parameters $\theta$ of the neural network $a_\theta(\cdot, \cdot)$:
\begin{equation}
    \alpha_{i,j} = \frac{\exp \left( a_\theta(\mathbf{x}_i(t), \mathbf{x}_j(t)) \right)}{\sum_{l \in \mathcal{N}_1(i) \cup \{ i \}} \exp \left( a_\theta(\mathbf{x}_i(t), \mathbf{x}_l(t)) \right)}.
\end{equation}

\section{Higher Order Graphical Attention}

While successfully applying attention to local connections, strictly edge-wise attention does not explicitly consider long-distance relationships. Instead, it relies on the reciprocation of messages via operators acting on local connections. Messages are subsequently lost via propagation through graph bottlenecks \cite{topping2021understanding}. Message-passing considering nodes of variable shortest path distance from each other possesses attractive theoretical properties, such as expressivity that surpasses the 1-WL isomorphism test \cite{zhang2023rethinking}. We formulate the Higher Order Attention (HoGA) module, as shown in Figure \ref{fig:walk_proc}, which samples the $k$-hop neighborhoods up to a maximal distance $K$ around a node of interest, constructing an attention matrix that describes higher order relationships within the graph. By capturing these relationships, our HoGA module improves the ability of MPNNs to recognize complex structural patterns.

\subsection{HoGA Formulation} 

HoGA directly considers the effect of node $j$ on node $i$ with shortest path distance $k$ via learning the impact of $\mathbf{x}_j(t)$ on $\mathbf{x}_i(t)$. We define the shortest path between $i$ and $j$ as $\mathcal{P} = (i=i_{1}, i_2,\dots, j=i_{k})$, and introduce a new attention coefficient, $\alpha_{i,j,k}$, which describes their order $k$ relationship, weighting $\mathcal{P}$.

Our attention module only considers one such path, as we are concerned with the endpoint feature-vector $\mathbf{x}_j(t)$, as opposed to the remaining topological substructure of the graph described by $\mathcal{P}$. In this sense, our approach contrasts with walk-based methods for multi-hop feature extraction \cite{michel2023path,kong2022geodesic,yeh2023random}, which suffer from sequential sampling bias, and must by necessity obtain a substantial number of walk  $\mathcal{P}$ for the precise reason of capturing the full topology of the graph \cite{yeh2023random}. HoGA, however, achieves this via $K$ parameterisations of a $k$-order line graph:

\noindent
\begin{definition}[$k$-Order Line Graph]
    The $k$-order line graph transform $L_k$ of a graph $G = (V, E)$ is a mapping that produces a new graph $L_k(G) = (V, E_k)$, where $(i, j) \in E_k$ if and only if there exists a shortest path of length $k$ between nodes $i$ and $j$ in $G$.
\end{definition}

Taking a walk on these $K$ line graphs allows for reduced bias to localized substructures encountered between endpoints of walk $\mathcal{P}$. In particular, following the strictly stronger expressivity of $k$-hop aggregation schemes than the $1$-WL isomorphism test \cite{zhang2023rethinking}, our model applies the weight $\alpha_{i,j,k}$ to paths of length $k$ varying such that $1 \leq k \leq K$. 

\begin{figure}[t]
    \centering
        \centering \includegraphics[width=0.8\columnwidth]{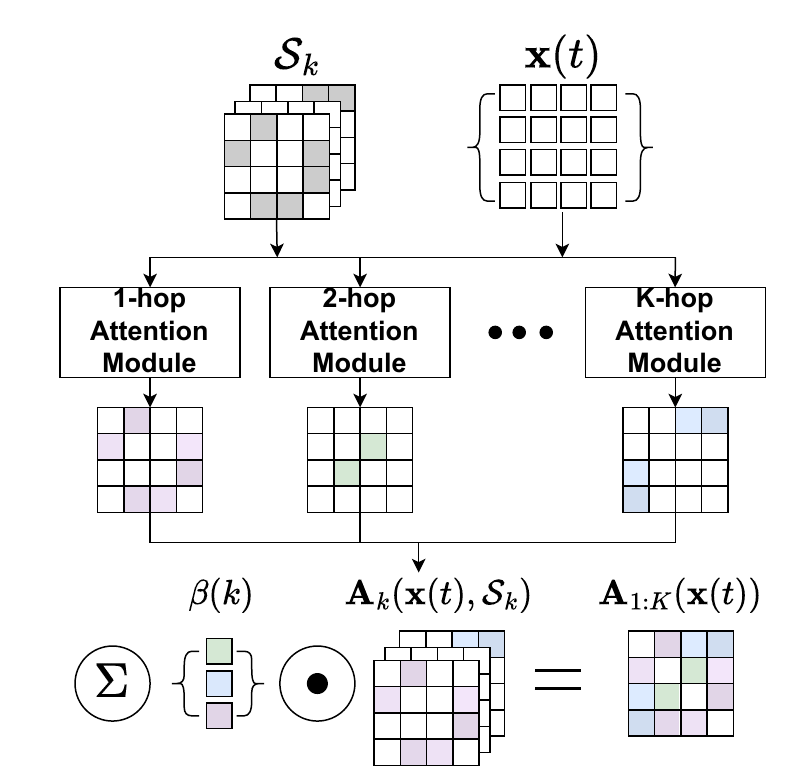}
        \caption{Our higher-order attention module aggregates weights from a single-hop self-attention method by weighting contributions proportional to proximity.}
        \label{fig:concept}
\end{figure}

To design a tractable parameterisation, we take a walk $\mathcal{S}_k \subset E_k$ on the $k$-order line graph $L_k(G) = (V, E_k)$ of $G$. Thereby, we induce a new adjacency matrix $\mathbf{A}_k(\mathbf{x}(t), \mathcal{S}_k)$ depending on the subset $\mathcal{S}_k$ for connectivity:
\begin{equation}
    \mathbf{A}_k(\mathbf{x}(t), \mathcal{S}_k)_{i, j} = 
    \begin{cases}
        \alpha_{i,j,k} & (i, j) \in \mathcal{S}_k, \\
        0 & \text{otherwise}.
    \end{cases}
\end{equation}
In this context, the attention coefficient $\alpha_{i,j,k}$ is computed via the neural network $a_{\theta_{k}}(\mathbf{x}_i(t), \mathbf{x}_j(t))$ with parameters $\theta_k$. The final representation, shown in Figure \ref{fig:concept}, is given by $\mathbf{A}_{1:K}(\mathbf{x}(t))$, where we have dropped dependence on $\mathcal{S}_k$ for ease of notation:
\begin{equation}
    \label{eq:multi_hop_att}
    \mathbf{A}_{1:K}(\mathbf{x}(t)) = \sum_{1 \leq k \leq K} \beta(k) \mathbf{A}_k(\mathbf{x}(t), \mathcal{S}_k). 
\end{equation}
Here $\mathcal{S}_1 = E$. The function $\beta: \mathbb{R} \mapsto \mathbb{R}$, first investigated by Wang et al. \cite{wang2020multi}, represents a weighting function that scales messages disproportionately to their commute times, simulating the actual transit times of the message while avoiding over-squashing problems. Introducing $\beta$ reduces the risk of overfitting to the long-distance signals during training. For simplicity, we consider $\beta(k)$ as the harmonic series $\beta(k) = \frac{1}{k}$.   

\subsection{Sampling the $k$-hop neighborhood} 

We define the branching factor $b$ of $G$ as the average node degree. A key problem in any direct parameterization of the shortest paths of length $k$ is that the $k$-hop neighborhood $\left\vert \mathcal{N}_k(i) \right\vert$ grows at order $O(b^k)$. Some works avoid this issue by simply aggregating the entirety of the periphery graph \cite{abboud2022shortest}. However, given the corresponding exponential growth in node diversity, this approach also reduces the utility of $\mathcal{N}_k(i)$ when $k$ increases, leading to a decrease in performance. A natural question arises: {\it how can a tractable parameterization of $\mathcal{N}_k(i)$ be constructed which simultaneously respects the diversity of the feature vectors and class labels?} 

We propose a sampling method that maximizes the diversity of node feature vectors in a subset $\mathcal{S} \subset \mathcal{N}_k(i)$. Our sampling methods furthermore ensure tractability by setting $\left \vert \mathcal{S} \right \vert = \left \vert E \right \vert$. 
We require our estimation of the $k$-hop feature space to converge adequately within $\vert E\vert$ steps. Additionally, we have constrained the number of parameters $dim(\theta_k)$, such that it grows linearly with the size of $G$. 
For $k$hops, we have an asymptotic growth of order $O(k\cdot\left\vert E \right\vert)$ for the number of non-zero entries in $\mathbf{A}_{1:K}(\mathbf{x}(t))$. 

{\bf Heuristic Probabilistic Walk.} We formulate a simple walk-based method that aims to  select $(i, j) \in E_k$, where the expected discrepancy between $\mathbf{x}_i = \mathbf{x}_i(0)$ and $\mathbf{x}_j = \mathbf{x}_j(0)$ is maximal. Given a history buffer $H = \{\!\{ \mathbf{x}_1, \dots, \mathbf{x}_n \}\!\}$ of size $n$ and a current node of interest $i \in V$, we select a node $j$ from candidate set $\mathcal{N}_k(i)$ with probability $p \sim s_n$, where $s_n$ is the dissimilarity score between feature vectors:
\begin{equation}
    s_n = \gamma \cdot f(\mathbf{x}_i, \mathbf{x}_j) + (1 - \gamma) \cdot f(\hat{\mathbf{x}}, \mathbf{x}_j),
    \label{eq:dissim}
\end{equation}
\begin{equation}
    f(\mathbf{x}_i, \mathbf{x}_j) = 1 - \frac{\mathbf{x}_i \cdot \mathbf{x}_j}{\|\mathbf{x}_i\| \cdot \|\mathbf{x}_j\|},
\end{equation}
\begin{equation}
    \hat{\mathbf{x}} = \sum_{1 \leq t \leq n} \gamma^{n-t} \cdot \mathbf{x}_{i_t}.
    \label{eq:buffer}
\end{equation}

 The cosine dissimilarity $f(\cdot, \cdot)$ measures the discrepancy between $\mathbf{x}_i$ and $\mathbf{x}_j$ in terms of their collinearity. The parameter $\gamma \in [0, 1]$ is the decay rate, and $\hat{\mathbf{x}}$ represents an exponential moving average over $H$. Intuitively, the first part of Equation \ref{eq:dissim} is a greedy step, as it tends to choose a $\mathbf{x}_j$ of maximal difference to $\mathbf{x}_i$, whereas the second part enforces a global dissimilarity for all visited nodes in $H$. The term $\gamma$ acts as the decay rate in Equation~\ref{eq:buffer}, and balances the contributions between the greedy and history buffer steps. Appendix {\bf A} discusses the preprocessing and runtime. 

\begin{figure}[t]
        \centering        \includegraphics[width=0.85\columnwidth]{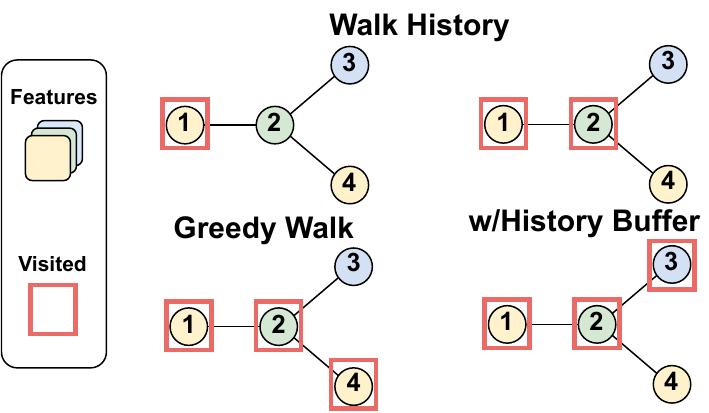}
        \caption{The history buffer stores concepts previously seen in the $k$-hop neighborhood to avoid repetitively resampling, allowing for greater capture of diverse higher-order relationships.}
        \label{fig:sample_fig}
\end{figure}
Figure~\ref{fig:sample_fig} illustrates the HiGFlow example and shows how it mitigates repetitive concept sampling from the state space of possible $k$-order relationships. When a concept is already present in the history buffer (highlighted in yellow), its sampling probability decreases. The following theorem and corollary formalize how HoGA reduces the likelihood of repetitive sampling by avoiding re-entry into cycles.

\begin{theorem}[Sampling Repetition on Cycles]
    Let $ C = (j_1, \dots, j_L) $ be any cycle of length $L$. The probability that a walk traverses the cycle exactly in order is: 
    \begin{equation}
        \mathbb{P}(i_n = j_1, \dots, i_{n+L} = j_L \mid C) = \prod_{1 \leq l \leq L} \frac{\omega_{j_l, j_{l+1}, n+l}}{\sum_{m \in \mathcal{N}_k(j_l)} \omega_{j_l, m, n+l}}.
    \end{equation}
    The edge weights $\omega_{i,q,\tau}$ are updated each walk iteration by the non-greedy component of the walk, and are given by:  
    \begin{equation}
        \omega_{i, q, \tau} = \gamma \left(f(x_i, x_q) - \delta_q(H_\tau)\right) + \delta_q(H_\tau),
    \end{equation}
    with $\delta_q(H_\tau)$ denoting the history buffer term at time $\tau$, and $f$ a dissimilarity function.
    \label{th:th1}
\end{theorem}

\begin{corollary}[History buffer on Cycles]
    Suppose the cycle $C = (j_1, \dots, j_L)$ is fully contained within the history buffer $H_\tau$. Then, the probability of avoiding the cycle is proportional to:
    \begin{equation}
        \frac{\delta_q(H_\tau)}{\sum_{m \in \mathcal{N}_k(i_{n+l-1})} \delta_m(H_\tau)},
    \end{equation}
    where $\delta_q(H_\tau)$ denotes the history buffer constraint at time $\tau$.
    \label{th:th2}
\end{corollary}

In contrast, a greedy walk re-enters cycles with constant probability, repeatedly sampling from the union of all cycles in the subgraph. Our corollary shows that, by leveraging a history buffer, the walk avoids previously visited topological substructures. Specifically, the history buffer constraint $\delta_q(H_\tau)$ acts as a repulsion mechanism, accelerating convergence of the empirical distribution to the true underlying distribution over $G$. Its effect is inversely proportional to the multiplicity of $q \in H_\tau$. We prove Theorem 1 and Corollary 1 in Appendix~\textbf{B}.

{\bf Higher Order Attention Heads.} For any sampling procedure, our method enables the computation of multiple attention heads via effectively allowing $i \in V$ to resample with replacement from $\mathcal{N}_k(i)$. Specifically, given the subset $\mathcal{S}_k \subset E_k$ sampled from a distribution $P(\mathcal{S}_k)$, we define multi-head attention on the higher-order layer as the expected adjacency matrix over $P(\mathcal{S}_k)$:
\begin{equation}
    \begin{split}
        \mathbf{A}_{k}(\mathbf{x}(t)) 
        &= \mathbb{E}_{\mathcal{S}_k \sim P(\mathcal{S}_k)}[\mathbf{A}_{k}(\mathbf{x}(t), S_k)]\\
        &\approx \frac{1}{\left\vert \Gamma \right\vert} \sum_{\mathcal{S}_k \in \Gamma} \mathbf{A}_k(\mathbf{x}(t), \mathcal{S}_k). 
    \end{split}
\end{equation}
Here, $\Gamma$ is a super-set containing samples from $P(\mathcal{S}_k)$. In the next section, this definition is consistent with simply taking the expectation over feature vectors $\mathbf{x}(t+1)$ for the models we consider. To preserve the correlation of feature vectors across layers, we fix $\mathcal{S}_k$ with respect to $t$, allowing the network depth to extract feature vectors at higher resolutions.

\subsection{Extending Single-hop Attention Models}

Attention forms the backbone of many graphical models \cite{velickovic2018graph,chamberlain2021grand,choi2023gread}. We demonstrate the versatility of HoGA by integrating the attention module into two distinct attention-based models. Specifically, in our research, we evaluate the application of the HoGA module to existing graphical models where attention is a fundamental component, explicitly focusing on GAT \cite{velickovic2018graph} and GRAND \cite{chamberlain2021grand}. In general, as shown in Figure \ref{fig:concept}, we replace their single-hop adjacency matrix $\mathbf{A}(\mathbf{x}(t))$ with $\mathbf{A}_{1:K}(\mathbf{x}(t))$.

We now summarize these single-hop attention models. The GAT model computes self-attention weights for each edge $(i, j) \in E$ at layer $t$. Specifically, node $i$ selects a subset of its neighboring edges $j \in \mathcal{N}_1(i)$ by calculating the attention matrix using Equation~\ref{eq:attn}:
\begin{equation}
    \mathbf{x}(t + 1) = \mathbf{A}(\mathbf{x}(t), t) \cdot \mathbf{x}(t).
\end{equation}
In our HoGA-GAT model, we generalize $\mathbf{A}(\mathbf{x}(t), t)$ to $\mathbf{A}_{1:K}(\mathbf{x}(t), t)$, as defined in Equation~\ref{eq:multi_hop_att}. We therefore introduce a multi-hop attention mechanism with an added dependency on the layer index $t$. 

GRAND belongs to the neural flow family of models \cite{bilovs2021neural}, which rely on the graph structure to describe message-passing as a physical process, {\it i.e.}, governed by a partial differential equation. For comparison, we use the GRAND model with Laplacian attention, where the adjacency matrix parameters are shared across all layers. The GRAND model is also expressed in terms of the attention matrix from Equation~\ref{eq:attn}.
\begin{equation}
    \label{eq:GRAND}
    \frac{\partial \mathbf{x}(t)}{\partial t} = \mathbf{A}(\mathbf{x}(t)) \cdot \mathbf{x}(t).
\end{equation}
Numerical methods, \textit{e.g.}, forward Euler, are used to solve Equation~\ref{eq:GRAND}. The network layer with index $t$ corresponds to the solution of Equation~\ref{eq:GRAND} at time $t$. In our HoGA-GRAND model, we replace the attention matrix $\mathbf{A}(\mathbf{x}(t))$ with its multi-hop formulation from Equation~\ref{eq:multi_hop_att}.

\begin{table*}[ht]
    \renewcommand{\arraystretch}{0.90}
    \caption{We compare attention-based models with and without our higher-order module. Best results are bold; second best are underlined. Superscript $^*$ marks the most significant model by the Wilcoxon signed-rank test, when applicable.}
    \begin{tabular}{lcccccccc}
        \toprule
         \textbf{Baselines} & \textbf{Cora} & \textbf{Citeseer} & \textbf{PubMed} & \textbf{Computers} & \textbf{Actor} & \textbf{Photo} & \textbf{Wisconsin} & \textbf{Texas} \\
        \midrule
        \multicolumn{1}{c}{} & \multicolumn{8}{c}{$1$st-Order Models}  \\ 
        \midrule 
        ChebNet \cite{defferrard2016convolutional}  & $79.8\pm0.5$ & $69.0\pm1.0$ & $77.9\pm0.4$ & $90.8\pm0.5$ & $51.3\pm1.1$ & $95.6\pm1.0$ & $79.6\pm2.0$ & $78.9\pm2.6$ \\ 
        GCN \cite{kipf2016semi}  & $81.8\pm0.6$ & $70.9\pm0.6$ & $78.3\pm0.4$ & $81.6\pm4.4$ & $53.5\pm1.2$ & $83.5\pm1.4$ & $56.5\pm1.5$ & $66.5\pm1.3$ \\
        GAT \cite{velickovic2018graph} & $81.6\pm0.9$ & $71.3\pm0.8$ & $77.0\pm1.0$ & $90.2\pm3.7$ & $40.6\pm0.9$ & $91.6\pm4.4$ & $55.1\pm3.8$ & $55.1\pm8.0$ \\ 
        JKNet \cite{xu2018representation}  & $79.0\pm1.3$ & $66.9\pm1.4$ & $76.0\pm0.8$ & $87.4\pm2.7$ & $53.2\pm0.8$ & $93.0\pm2.9$ & $58.4\pm0.8$ & $67.8\pm0.6$ \\ 
        APPNP \cite{gasteiger2018predict}  & $82.2\pm0.9$ & $69.9\pm0.8$ & $78.2\pm0.3$ & $88.9\pm0.9$ & $45.6\pm0.7$ & $94.0\pm0.4$ & $56.1\pm3.4$ & $64.8\pm1.7$ \\ 
        GPRGNN \cite{chien2020adaptive}  & $82.6\pm0.9$ & $69.4\pm1.4$ & $78.3\pm0.5$ & $87.9\pm0.8$ & $52.0\pm0.7$ & $93.1\pm0.8$ & $79.4\pm2.9$ & $75.7\pm3.1$ \\ 
        BernNet \cite{he2021bernnet}   & $73.2\pm1.6$ & $67.5\pm1.5$ & $73.3\pm1.2$ & $88.2\pm0.5$ & $51.9\pm0.8$ & $95.4\pm0.8$ & $78.8\pm2.9$ & $79.3\pm3.2$ \\ 
        GRAND \cite{chamberlain2021grand}   & $83.0 \pm 1.0$ & $70.2\pm1.2$ & $78.8\pm0.8$ & $85.2\pm1.2$ & $41.2\pm0.9$ & $95.5\pm0.3$ & $75.4\pm1.0$ & $79.7\pm1.6$ \\
        DIFFformer \cite{wu2023difformer}  & $80.8\pm0.9$ & $70.9\pm1.3$ & $78.0\pm0.4$ & \hspace{3.8pt}$\best{94.5 \pm 0.4^{*}}$ & $46.1\pm0.8$ & $\underline{96.7 \pm 0.1}$ & $72.5\pm2.8$ & $80.5\pm3.1$ \\
        \midrule
        \multicolumn{1}{c}{} & \multicolumn{8}{c}{Higher-Order Models} \\
        \midrule
        MixHop \cite{abu2019mixhop}  & $81.6\pm0.8$ & $70.4\pm0.6$ & $78.8\pm0.8$ & $91.3\pm0.3$ & $\underline{59.2 \pm 0.5}$ & $92.1\pm0.3$ & $80.1\pm2.6$ & $81.1\pm1.7$ \\
        SPAGAN \cite{yang2021spagan} & $82.2\pm0.5$ & $\underline{72.4 \pm 0.7}$ & $77.9\pm0.6$ & $90.1\pm0.3$ & $31.3\pm0.6$ & $94.2\pm0.3$ & $51.4\pm5.0$ & $63.2\pm3.6$ \\
        STAGNN \cite{huang2023tailoring}   & $82.4\pm0.9$ & $71.0\pm0.8$ & $77.5\pm0.6$ & $90.4\pm0.4$ & $43.5\pm0.3$ & $94.6\pm0.3$ & $78.9\pm3.2$ & $80.8\pm2.3$ \\
        HiGCN \cite{huang2024higher} & $\underline{83.5 \pm 0.6}$ & $71.5\pm1.0$ & $\underline{79.4 \pm 0.5}$ & $92.2\pm0.6$ & $48.8\pm0.6$ & $96.6\pm0.2$ & $\underline{80.9 \pm 2.8}$ & $\underline{81.4 \pm 1.5}$ \\
        HoGA-GAT & $82.5\pm0.7$ & \hspace{3.8pt}$\best{73.0 \pm 0.4^{*}}$ & $78.3\pm0.4$ & $\underline{93.0 \pm 0.5}$ & \hspace{3.8pt}$\best{60.6 \pm 1.6^{*}}$ & $96.3\pm2.0$ & $60.0\pm1.0$ & $73.3\pm2.0$ \\
        HoGA-GRAND & \hspace{3.8pt}$\best{84.1 \pm 0.3^{*}}$ & $71.1\pm1.2$ & \hspace{3.8pt}$\best{80.6 \pm 0.5^{*}}$ & $92.8\pm1.9$ & $48.3\pm1.0$ & \hspace{3.8pt}$\best{98.1 \pm 1.7^{*}}$ & $\best{81.7 \pm 1.0}$ & \hspace{3.8pt}$\best{83.2 \pm 1.1^{*}}$ \\
        \bottomrule
    \end{tabular}
    \label{tab:results}
\end{table*}
\begin{table*}
    \renewcommand{\arraystretch}{0.90}
    \caption{Runtime (s) and GPU memory usage (GB) for Cora, Citeseer, and PubMed datasets. Values shown as mean $\pm$ standard deviation across $20$ runs.}
    \begin{tabular}{lrrrrrr}
        \toprule
        \textbf{Baselines} &
        \multicolumn{2}{c}{\textbf{Cora}} &
        \multicolumn{2}{c}{\textbf{Citeseer}} &
        \multicolumn{2}{c}{\textbf{PubMed}} \\
        \cmidrule(lr){2-3} \cmidrule(lr){4-5} \cmidrule(lr){6-7}
        & Runtime (s) & Memory (GB) & Runtime (s) & Memory (GB) & Runtime (s) & Memory (GB) \\
        \midrule 
        GAT \cite{velickovic2018graph} & $5.07\pm0.17$ & $0.04\pm0.00$ & $5.38\pm0.28$ & $0.79\pm0.02$ & $6.38\pm0.45$ & $0.62\pm0.00$ \\
        JKNet \cite{xu2018representation} & $5.62\pm0.11$ & $0.64\pm0.01$ & $5.34\pm0.14$ & $0.79\pm0.02$ & $6.89\pm0.47$ & $0.62\pm0.00$ \\
        APPNP \cite{gasteiger2018predict} & $5.28\pm0.10$ & $0.44\pm0.00$ & $5.09\pm0.22$ & $0.79\pm0.01$ & $5.53\pm0.47$ & $0.62\pm0.00$ \\
        GRAND \cite{chamberlain2021grand} & $0.82\pm0.29$ & $0.49\pm0.14$ & $1.05\pm0.29$ & $0.57\pm0.01$ & $2.13\pm0.32$ & $0.62\pm0.00$ \\
        \midrule 
        HoGA-GRAND (ours) & $3.46\pm0.22$ & $1.34\pm0.39$ & $4.22\pm0.28$ & $1.25\pm0.04$ & $9.28\pm0.28$ & $1.03\pm0.00$ \\
        HoGA-GAT (ours) & $8.38\pm0.34$ & $0.44\pm0.46$ & $8.77\pm0.24$ & $0.63\pm0.00$ & $8.50\pm0.40$ & $0.70\pm0.00$ \\
        \bottomrule
    \end{tabular}
    \label{tab:runtime-memory-cora-citeseer-pubmed}
\end{table*}

\section{Model Evaluation}

We conduct empirical studies to address the following question: \textit{Is the application of higher-order attention via direct sampling of the $k$-hop neighborhood a viable approach for multi-hop aggregation?} Our analysis is broken up into three components:
\begin{itemize}
    \item {\bf RQ1.} How do our higher-order models, HoGA-GAT and HoGA-GRAND, perform compared to existing higher-order techniques?
    \item {\bf RQ2.} To what extent does the HoGA model mitigate the oversmoothing effect, and is the model's accuracy stable across different choices of maximum nodal distance?
    \item {\bf RQ3.} What is the effectiveness of the heuristic walk sampling method compared to simpler sampling methods that do not consider global graph structure?
\end{itemize}
Specifically, we demonstrate the efficacy of our method through comparisons with state-of-the-art higher-order attention methods \cite{huang2024higher,abu2019mixhop,yang2021spagan}, and Fourier methods \cite{he2021bernnet,kipf2016semi,defferrard2016convolutional}.

\noindent{\bf Datasets.} We evaluate all models on core benchmark node classification datasets, wherein the dataset comprises a single graph: Cora, Citeseer, and Pubmed \cite{yang2016revisiting}. We also evaluate other diverse datasets of variable size: Amazon Computers, Amazon Photos, and CoAuthor Computer Science \cite{shchur2018pitfalls}. In addition to these homophilic datasets, we also evaluate on the heterophilic datasets, Wisconsin and Texas. 

{\bf Baseline Models.} We evaluate models that incorporate our higher-order attention module and compare them with other non-local aggregation schemes, which utilize either meta-paths, SPAGAN \cite{yang2021spagan}, and topological structure, HiGCN \cite{huang2024higher}. We also evaluate against STAGNN \cite{huang2023tailoring}, a multi-hop transformer model, and DIFFformer \cite{wu2023difformer}, a diffusion-based architecture. Furthermore, our study includes single-hop spectral methods such as APPNP \cite{gutteridge2023drew}, BernNet \cite{he2021bernnet}, and GCN \cite{kipf2016semi}. Additionally, we compare our higher-order attention models with their single-hop counterparts; GAT \cite{velickovic2018graph} and GRAND \cite{chamberlain2021grand}. We use the original parameter configurations for all baselines. 

\subsection{Reproducibility} 

We cover in this section various settings used throughout our empirical evaluations. The source code, along with data splits and samples of the $k$-hop neighborhood for all experiments, is provided in the supplementary materials. 

\noindent{\bf Data splits.} On Cora, Citeseer, and Pubmed, we use the public train, test, and validation set splits proposed in the original paper \cite{yang2016revisiting}. On the remaining datasets, we split the graph by randomly selecting nodes, where each set comprises 60\%, 20\%, and 20\% of all nodes.  

\noindent{\bf Higher-order sampling.} We set the random jump probability to 5\% when running our heuristic walk algorithm, and limit the maximum number of edges obtained from any sampling procedure to $90,000$ to reduce runtime. We use eight higher-order attention heads in the first network layer, and one in the subsequent layers. To avoid excessive fine-tuning, we set the maximum $k$-hop value to $K=3$ for all graphs, and keep the $k$-hop samples consistent across all layers. 

\noindent{\bf Experiment setup.} To reduce the variability in model performance due to random seed, we repeat all experiments $20$ times, re-initializing our models with a new seed at each repetition. To train and evaluate our models, we run our experiments using an A100 GPU. We evaluate the significance of the empirical model performance by using the Wilcoxon signed rank test with a confidence threshold of 5\% 

\definecolor{darkgreen}{rgb}{0.0, 0.5, 0.0}
\definecolor{darkblue}{rgb}{0.0, 0.0, 0.5}


\subsection{Experiments}

\begin{table*}[ht]
    \renewcommand{\arraystretch}{0.90}
    \caption{Ablation study on $k$-hop neighborhood sampling methods for HoGA-GAT. Bold indicates the highest accuracy, whereas underlined results indicates the second highest. The outcome of the Wilcoxon Signed Rank test is denoted by * given significance.}
    \begin{tabular}{lcccccc}
        \toprule
        \textbf{Samplers} & \textbf{Cora} & \textbf{Citeseer} & \textbf{PubMed} & \textbf{Computer} & \textbf{Photo} & \textbf{Actor} \\
        \midrule
        Random Sample & $81.4\pm0.6$ & $\underline{70.7\pm0.5}$ & $77.6\pm0.6$ & $88.2\pm1.4$ & $93.2\pm0.5$ & $32.7\pm0.7$ \\
        Random Walk   & $81.1\pm0.8$ & $70.5\pm0.7$           & $\underline{77.8\pm0.5}$ & $88.7\pm0.8$ & $93.5\pm0.8$ & $33.0\pm0.6$ \\
        Breadth First & $81.6\pm0.9$ & $69.6\pm0.8$           & $76.8\pm0.6$ & $84.8\pm1.3$ & $93.3\pm0.7$ & $32.5\pm0.6$ \\
        Depth First   & $81.3\pm0.9$ & $69.5\pm0.8$           & $77.0\pm0.9$ & $84.5\pm0.6$ & $93.2\pm0.5$ & $32.5\pm0.7$ \\
        Greedy        & $\underline{81.8\pm0.4}$ & $70.5\pm1.0$ & $\underline{77.8\pm0.5}$ & $\underline{91.8\pm0.6}$ & $\underline{95.3\pm0.2}$ & $\underline{33.6\pm0.3}$ \\
        Heuristic Walk & \hspace{3.8pt}${\bf 82.5}\pm{\bf 0.7^*}$ & \hspace{3.8pt}${\bf 73.0}\pm{\bf 0.4^*}$ & \hspace{3.8pt}${\bf 78.3}\pm{\bf 0.4^*}$ & \hspace{3.8pt}${\bf 93.0\pm0.5^*}$ & \hspace{3.8pt}${\bf 96.3\pm2.0^*}$ & \hspace{3.8pt}${\bf 60.6\pm1.6^*}$ \\
        \bottomrule
    \end{tabular}
    \label{tab:node_sampling_table}
\end{table*}
\begin{figure*}[ht]
    \centering
    \includegraphics[width=0.235\textwidth, height=24mm]{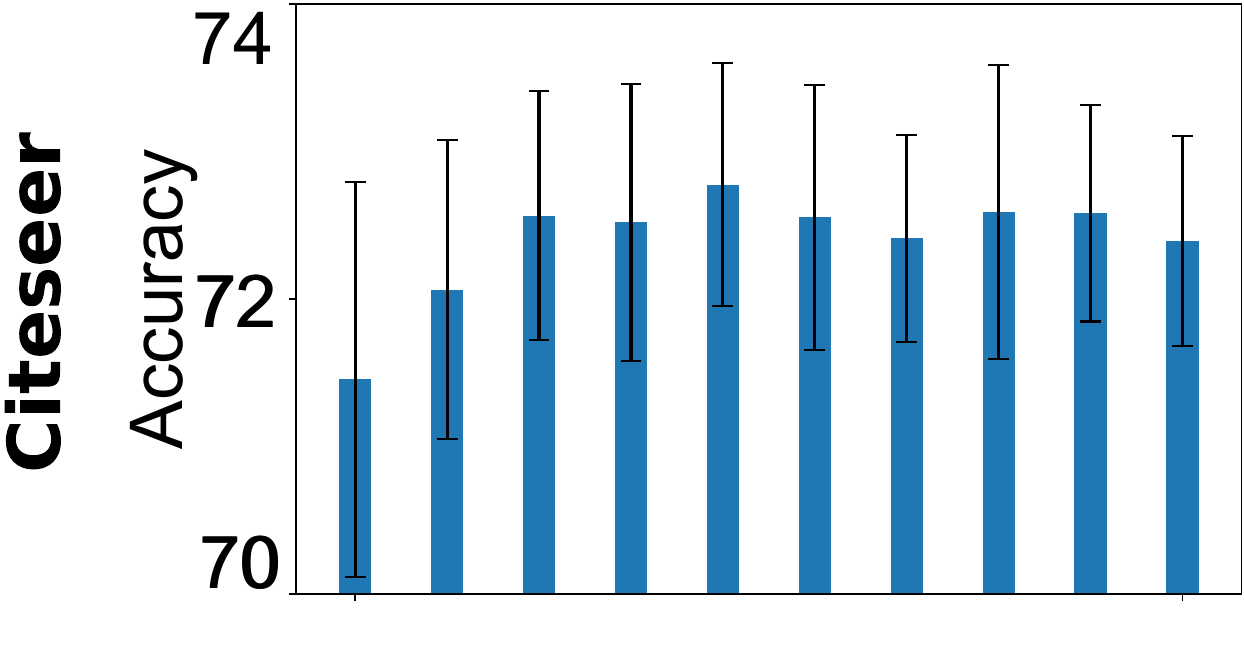}
    \hfill
    \includegraphics[width=0.21\textwidth, height=25.5mm]{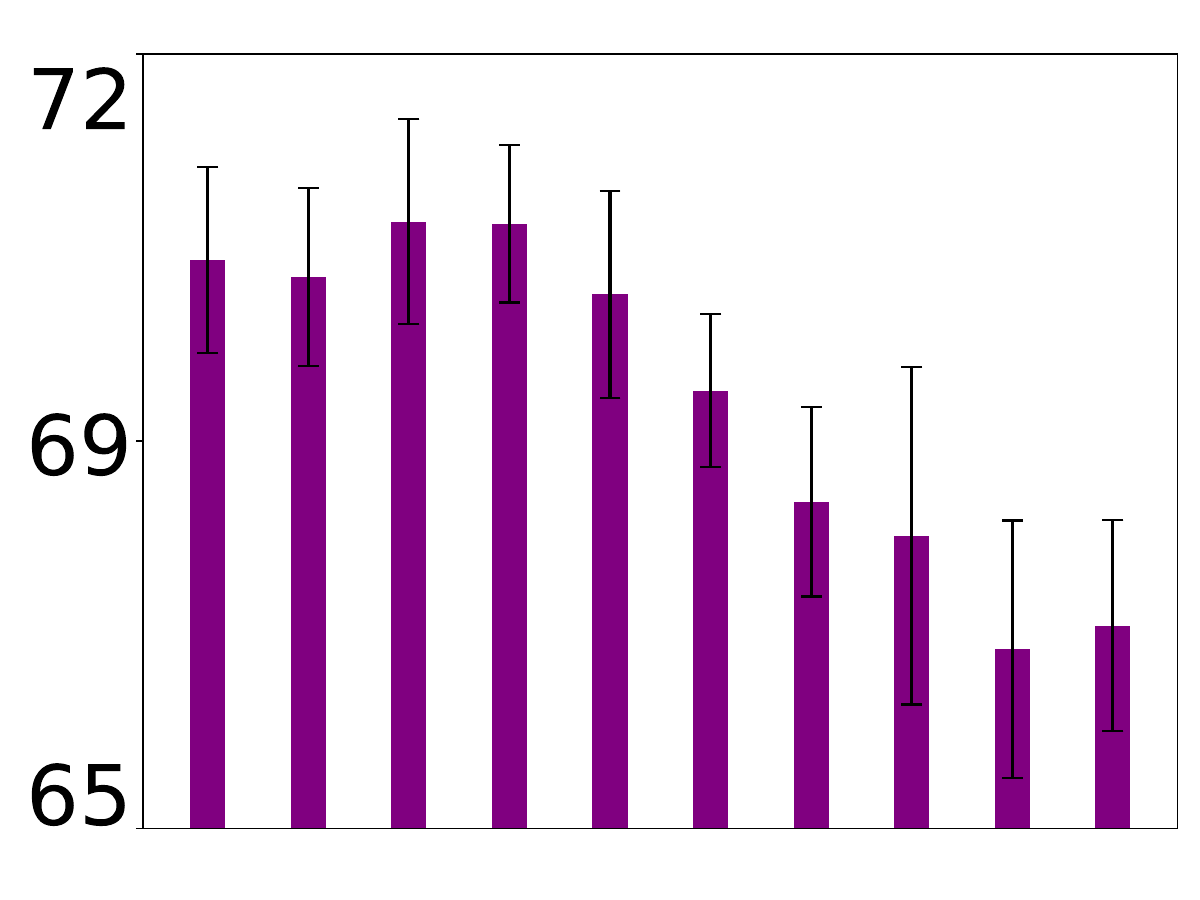}
    \hfill
    \hspace{-3.5mm}\raisebox{1.2mm}{\includegraphics[width=0.235\textwidth, height=23mm]{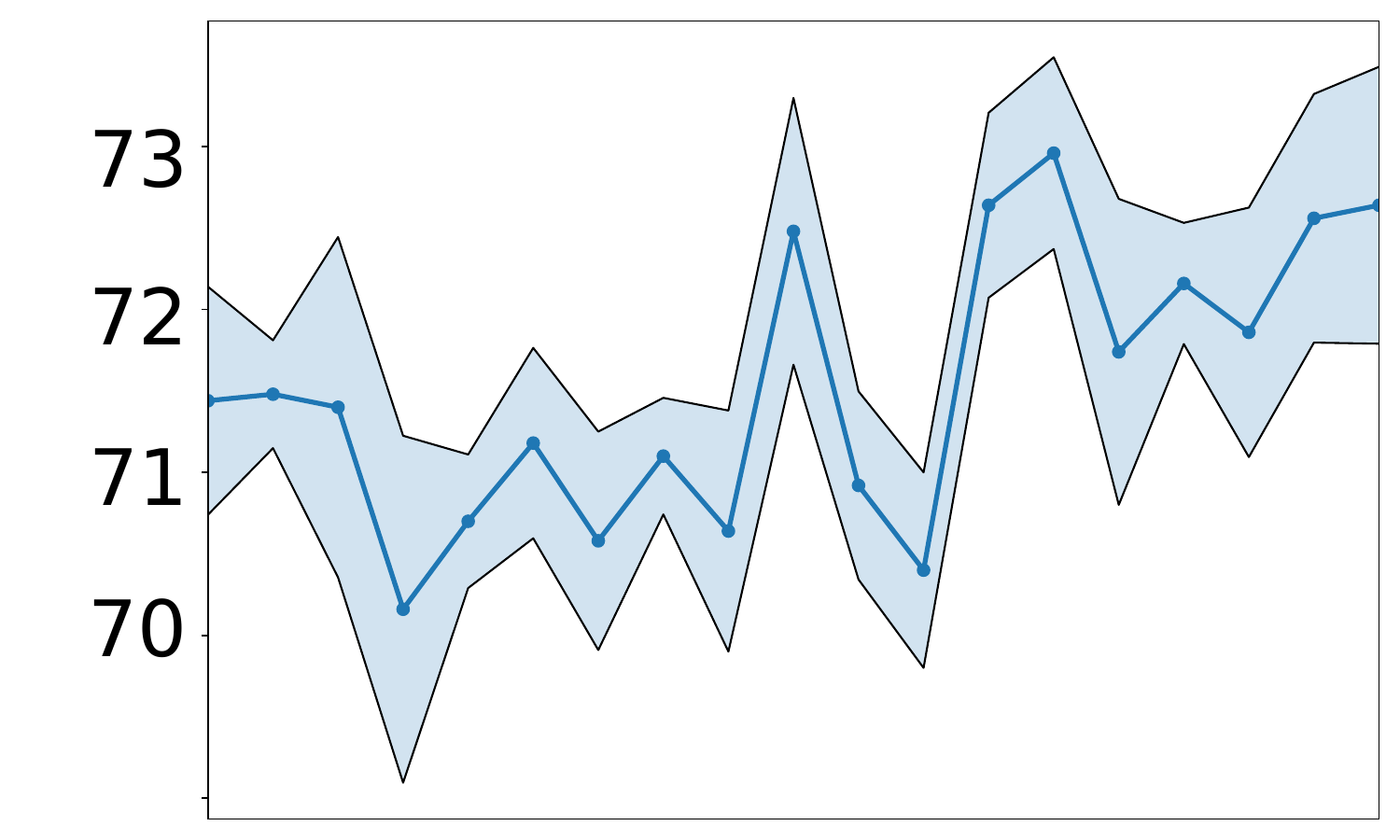}}
    \hfill
    \raisebox{-1mm}{\includegraphics[width=0.25\textwidth]{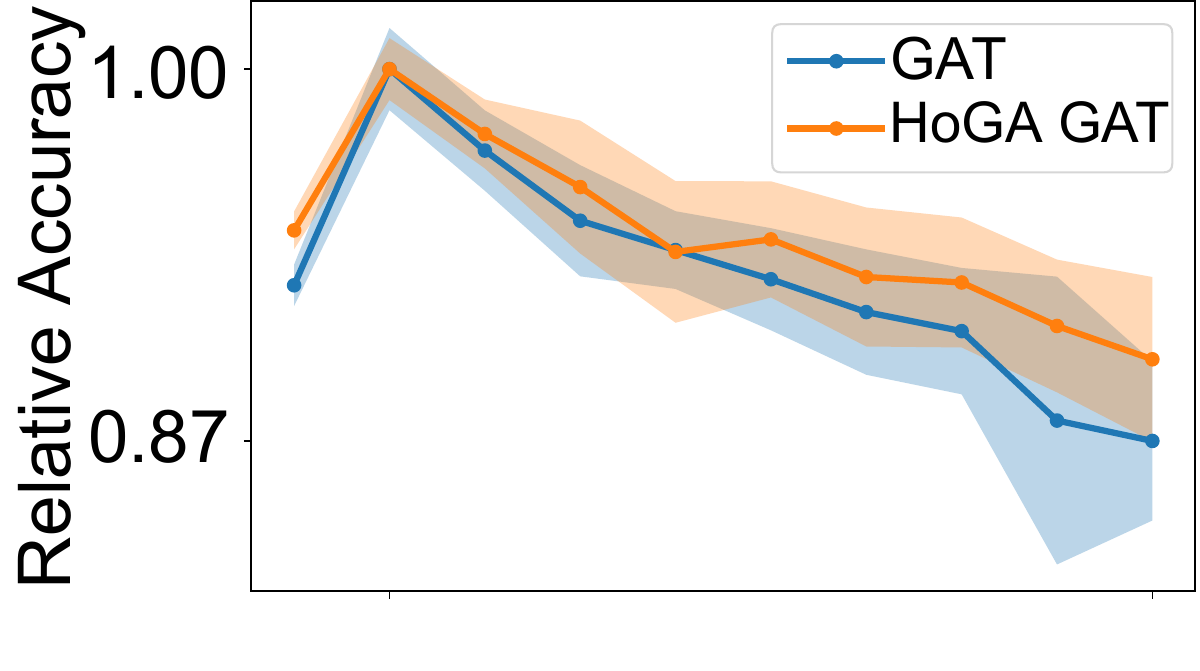}}\\

    \includegraphics[width=0.235\textwidth, height=23.5mm]{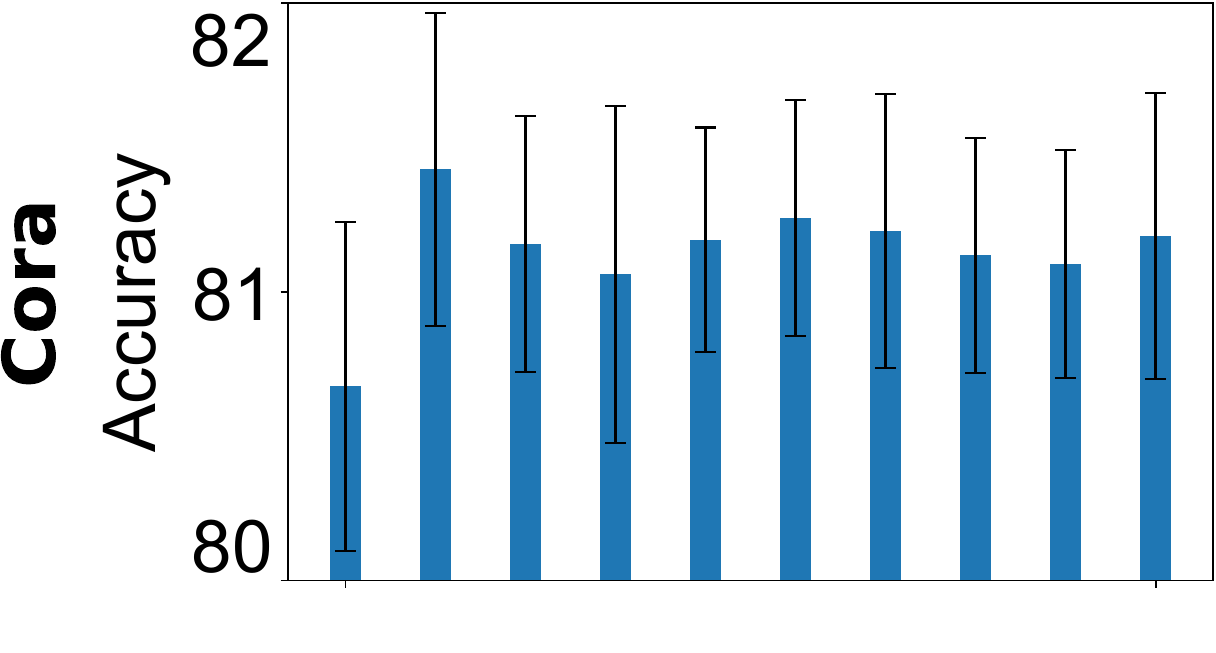}
    \hfill
    \includegraphics[width=0.21\textwidth, height=25mm]{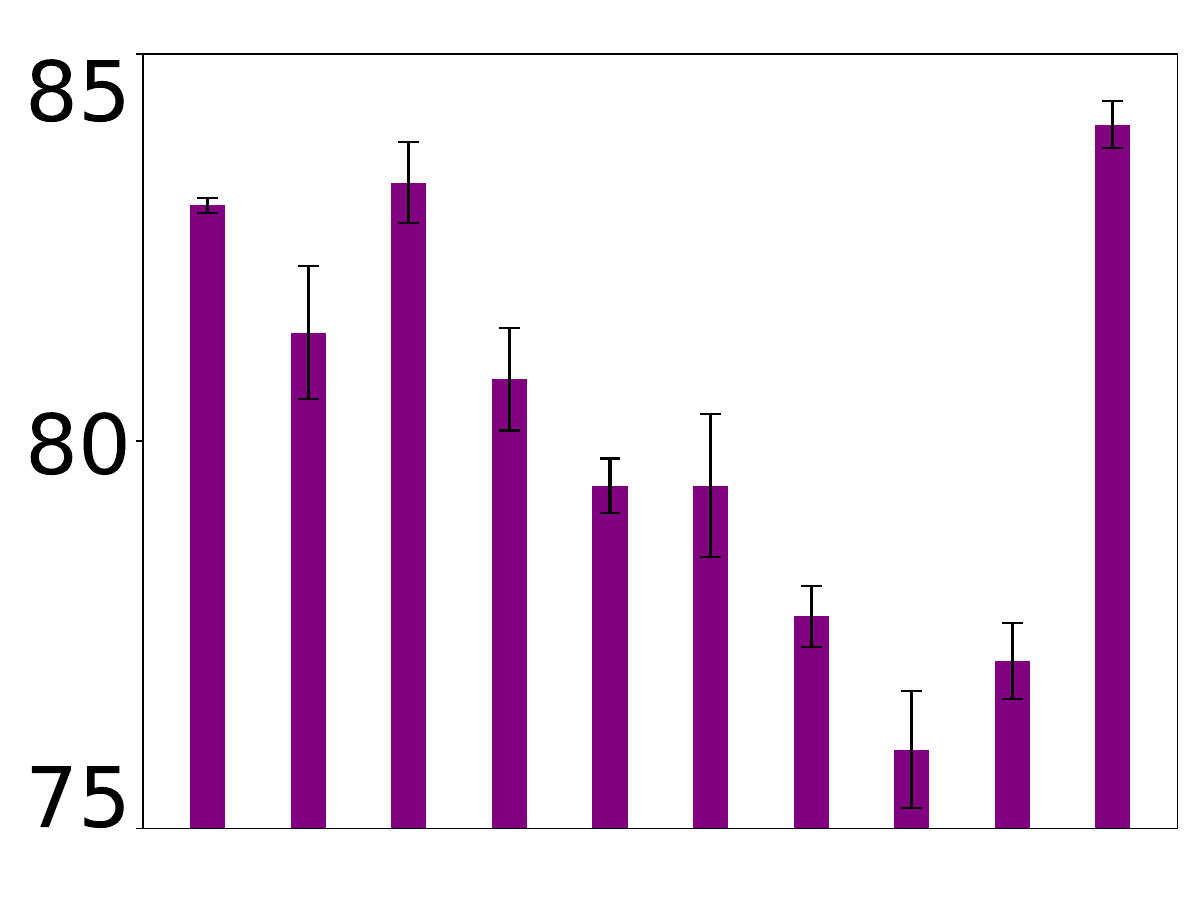}
    \hfill
    \hspace{-5mm}\raisebox{0.8mm}{\includegraphics[width=0.247\textwidth, height=23mm]{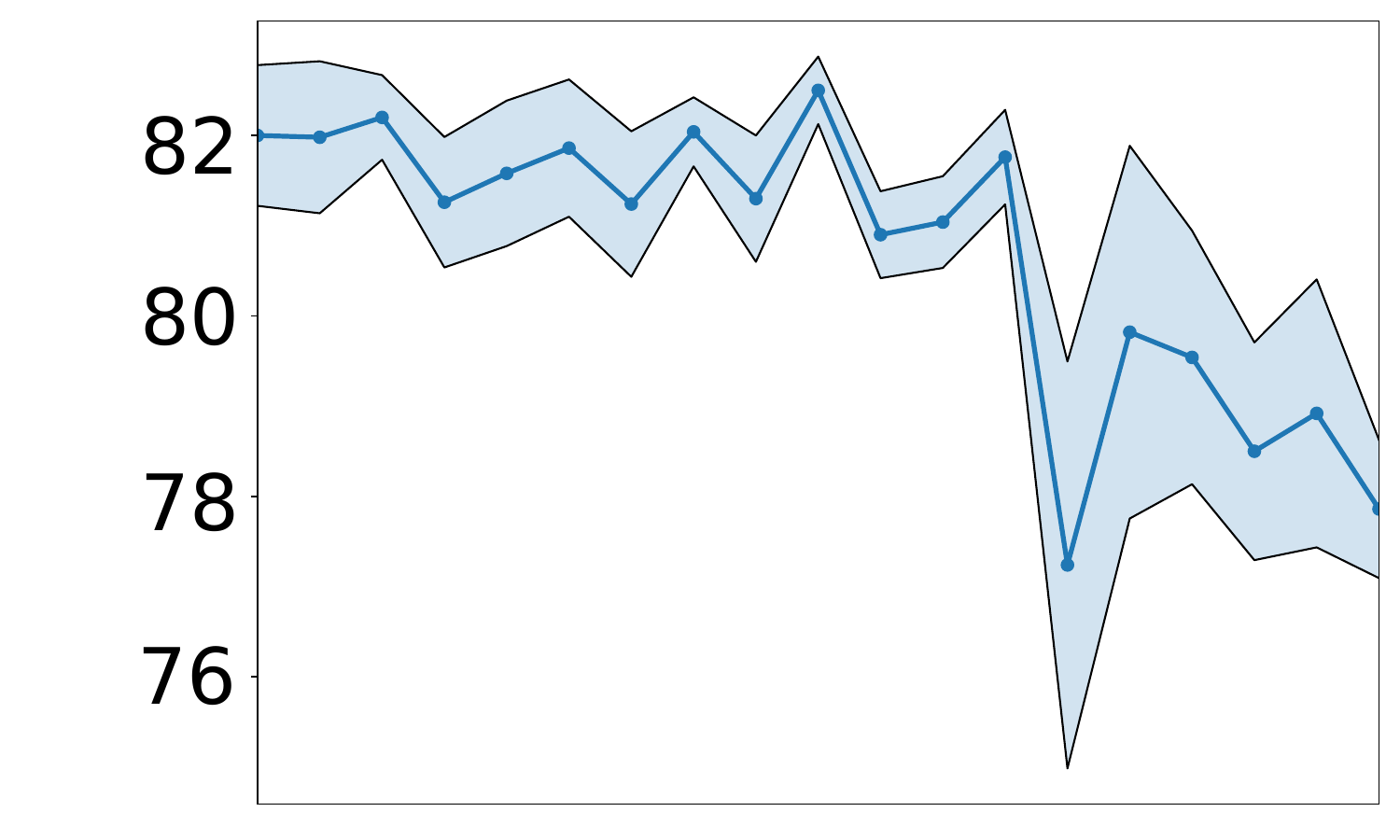}}\hspace{0.5mm}
    \hfill
    \raisebox{-1.3mm}{\includegraphics[width=0.25\textwidth]{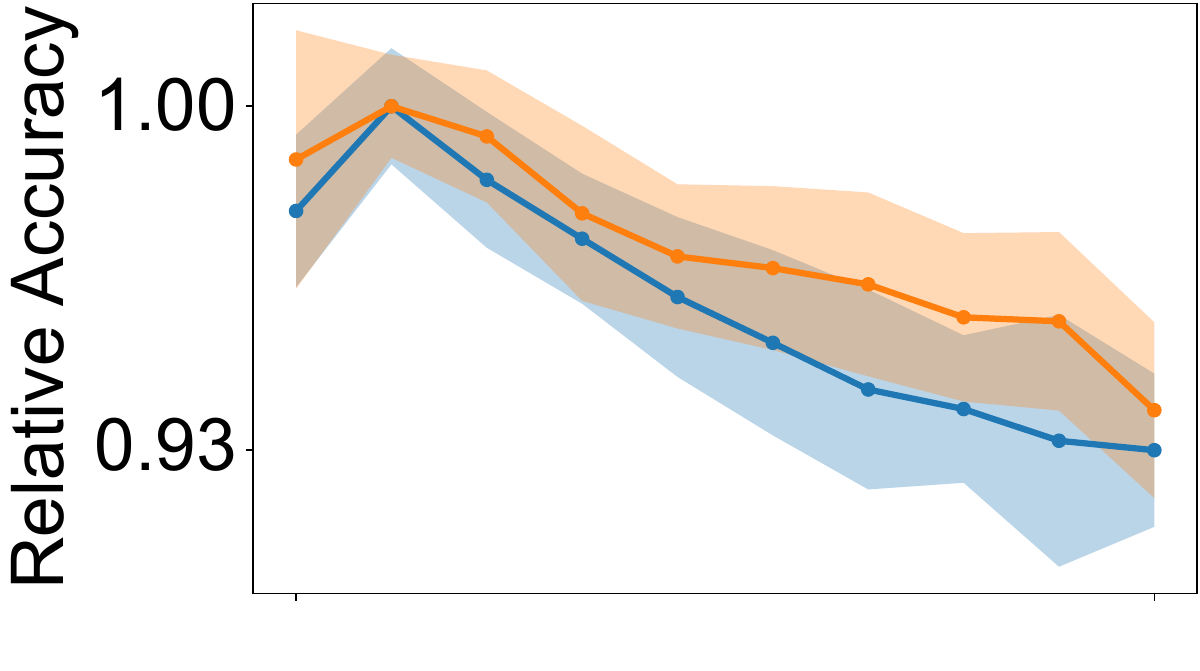}}\\

    \raisebox{1.3mm}{\includegraphics[width=0.235\textwidth, height=28mm]{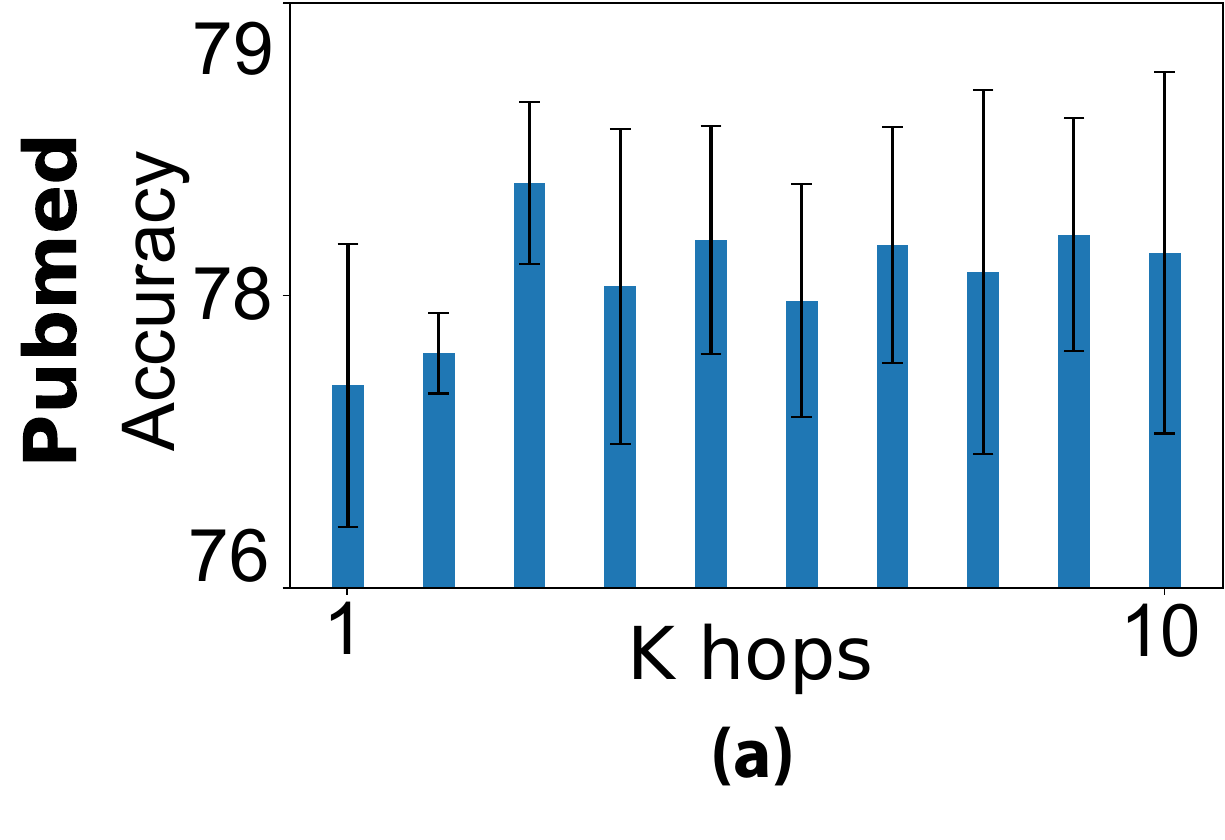}}
    \hfill
    \hspace{2mm}\raisebox{2.5mm}{\includegraphics[width=0.205\textwidth, height=27mm]{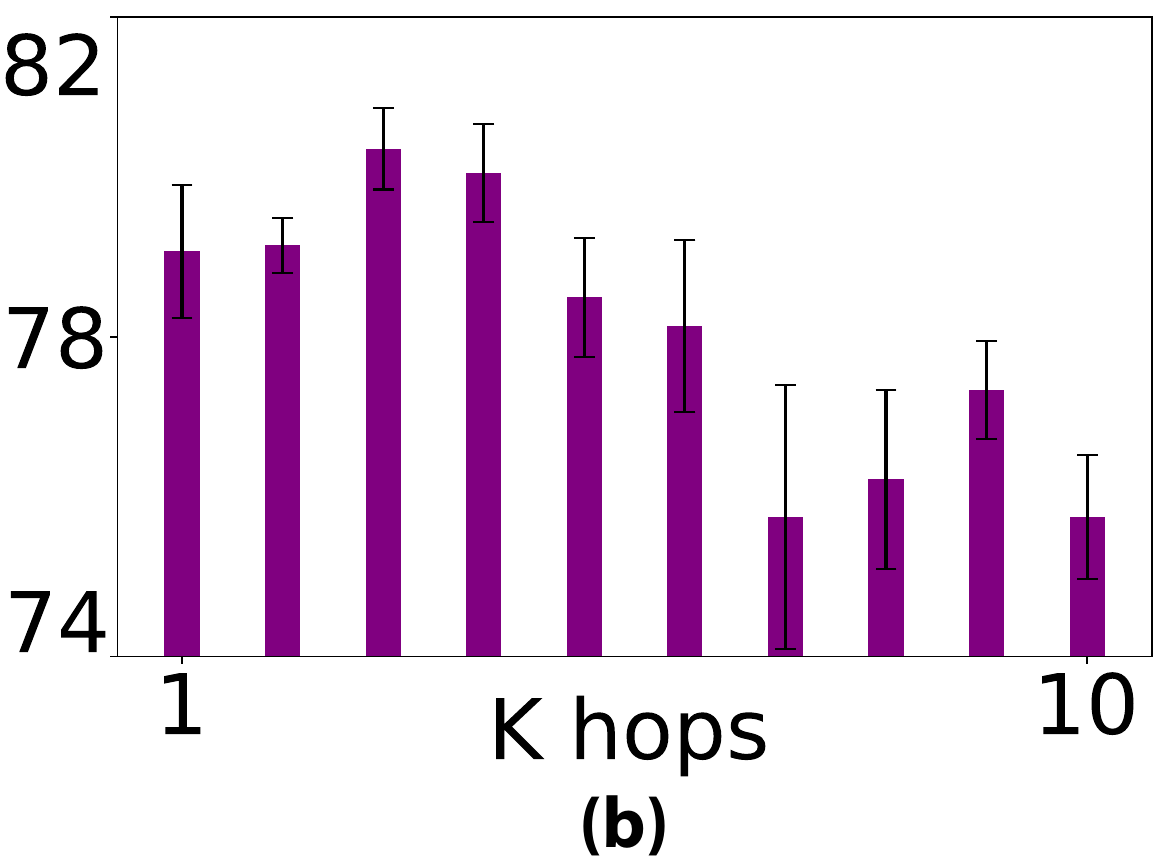}}
    \hfill
    \raisebox{3mm}{\includegraphics[width=0.24\textwidth, height=26.7mm]{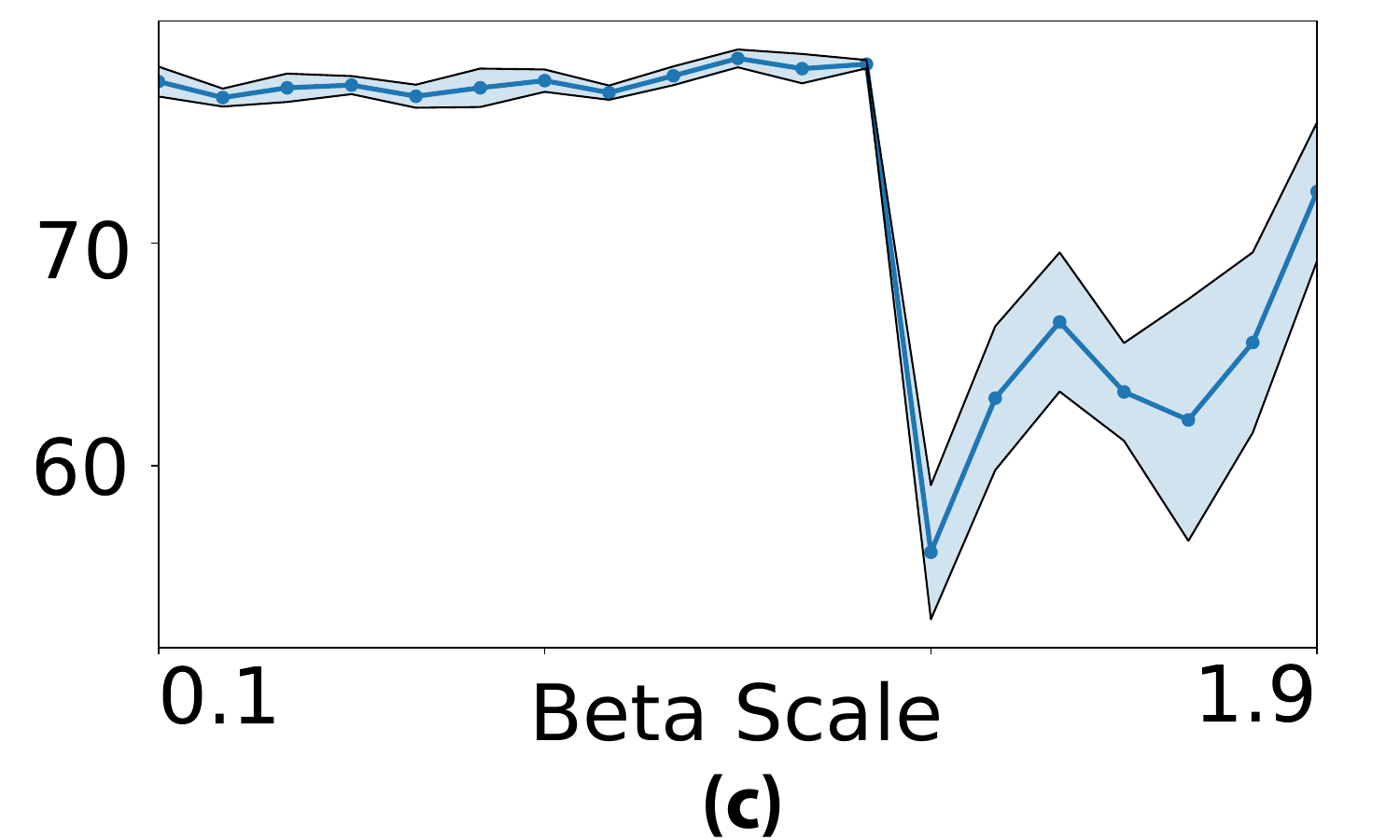}}
    \hfill
    \raisebox{0.3mm}{\includegraphics[width=0.25\textwidth, height=29mm]{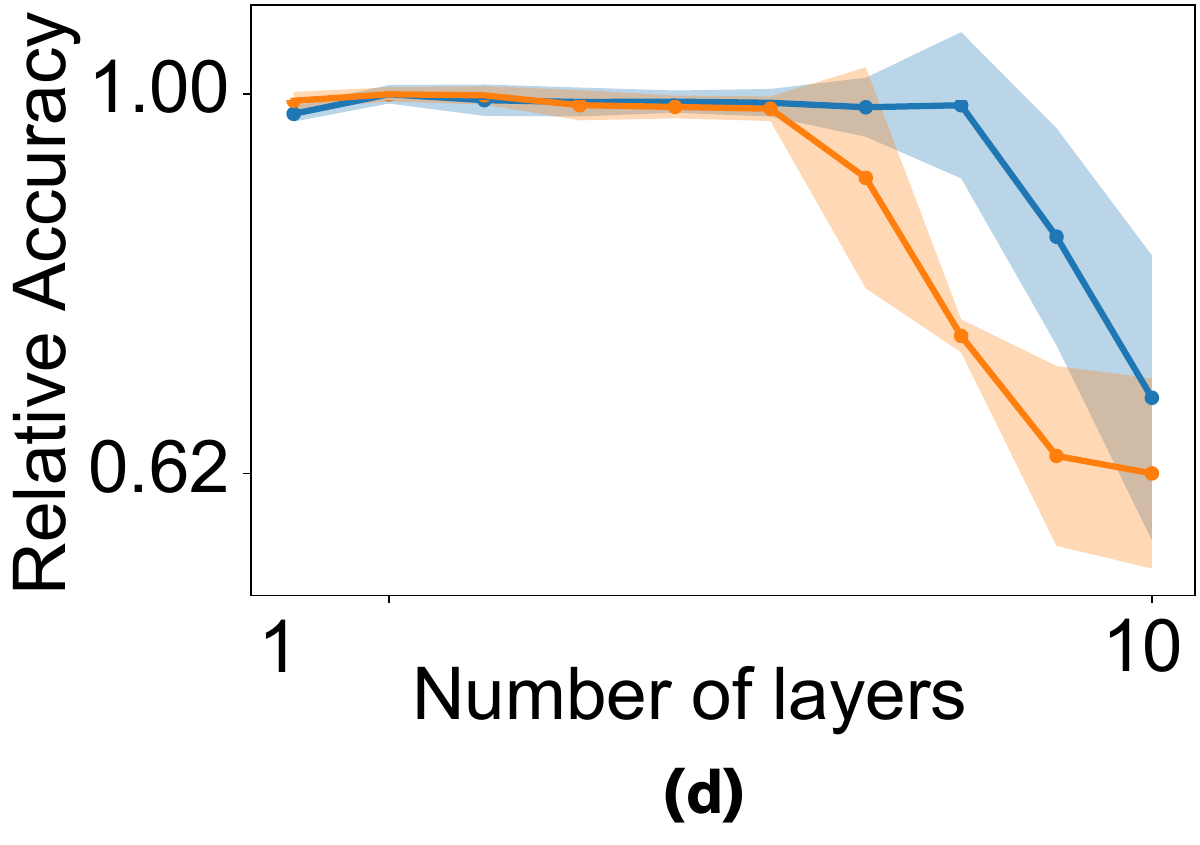}}

    \caption{Sensitivity tests with standard deviations across 20 iterations; {\bf (a)}-{\bf (b)} Varying maximal hop number for both HoGA-GAT and HoGA-GRAND models respectively, {\bf (c)} scaling factor multiplying $\beta(k)$, {\bf (d)} Relative accuracy under a variable number of layers.}
    \label{fig:oversmooth_ab_test}
\end{figure*}

We evaluate the efficacy of our HoGA module using node classification tasks as a proxy within our experimental setup. Our analysis compares accuracy with baseline models on benchmark datasets. We also perform qualitative assessments through sensitivity tests, focusing on the number of message-passing layers, maximum hop value $K$, and $\beta(k)$ regime.

{\bf Evaluation on Benchmark Datasets (RQ1).} Table~\ref{tab:results} presents the results of our experiments. Incorporating our attention module leads to an accuracy improvement ranging from 1.5\% on PubMed to 20\% on the Actors dataset, with an average increase of approximately 3\% across other benchmark datasets. These findings demonstrate that leveraging high-order information enables HoGA to achieve higher accuracy compared to the original single-hop model.

Overall, except on the Wisconsin dataset, either HoGA-GAT or HoGA-GRAND consistently achieves significantly higher accuracy compared to non-HoGA models. Table \ref{tab:results} specifically compares HoGA-GAT and HoGA-GRAND with HiGCN, a state-of-the-art higher-order attention model that maps topological substructures to similarity scores. For HoGA-GAT, while HiGCN outperforms it on Cora and Pubmed, HoGA-GAT demonstrates significantly higher accuracy on Citeseer, Computers, and Actor. The higher accuracy highlights that our higher-order attention paradigm, which evaluates similarity through feature vectors, provides superior performance over recent topology-based higher-order attention methods \cite{huang2024higher,zhang2024hongat}. In Wisconsin and Texas, however, GAT-based models, SPA-GAN, GAT, and HoGA-GAT, do not achieve high accuracy compared to other baselines. When comparing our attention module to SPAGAN, a meta-path sampling higher-order attention method, HoGA models generally deliver the best accuracy by a significant margin, except HoGA-GAT on Cora and HoGA-GRAND on Citeseer.

Intuitively, larger graphs exhibit greater modality in their feature-vector distributions and more intricate topological substructures in their $k$-order line graphs. Despite this, HoGA demonstrates significant accuracy improvements even on larger graphs. For example, HoGA-GAT achieves a 20\% and 5\% gain on Actor and Photo, respectively, compared to a 0.9\% gain on the smaller Cora graph. This indicates that the heuristic walk process effectively converges to a qualitatively accurate subgraph representation.

{\bf Runtime and Memory Comparison (RQ1).} We compare the peak GPU memory usage per training run, and the average runtime per epoch of our HoGA models against various baselines. We observe that while there is overhead from applying the HoGA model to both GAT and GRAND, this remains limited in terms of both GPU usage and runtime per epoch. This ensures that HoGA remains applicable. Furthermore, other baselines, such as JKNet and APPNP, are comparable to the HoGA models.

\begin{figure*}[t]
    \centering

    \begin{minipage}[t]{0.28\linewidth}
        \centering
        \includegraphics[width=\linewidth]{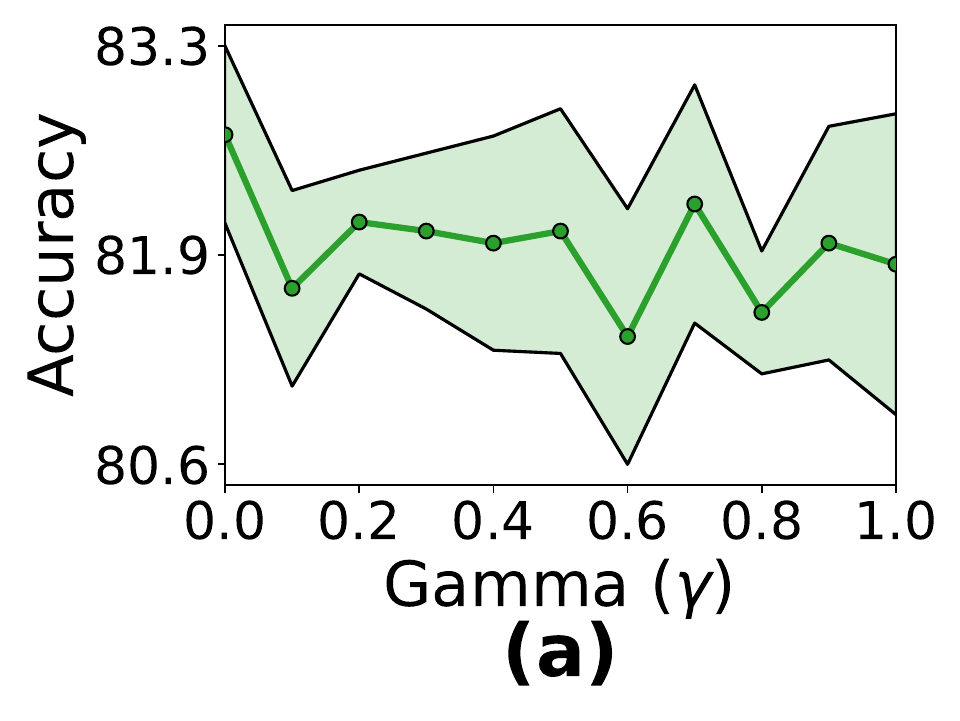}
    \end{minipage}
    \begin{minipage}[t]{0.28\linewidth}
        \centering\includegraphics[width=\linewidth]{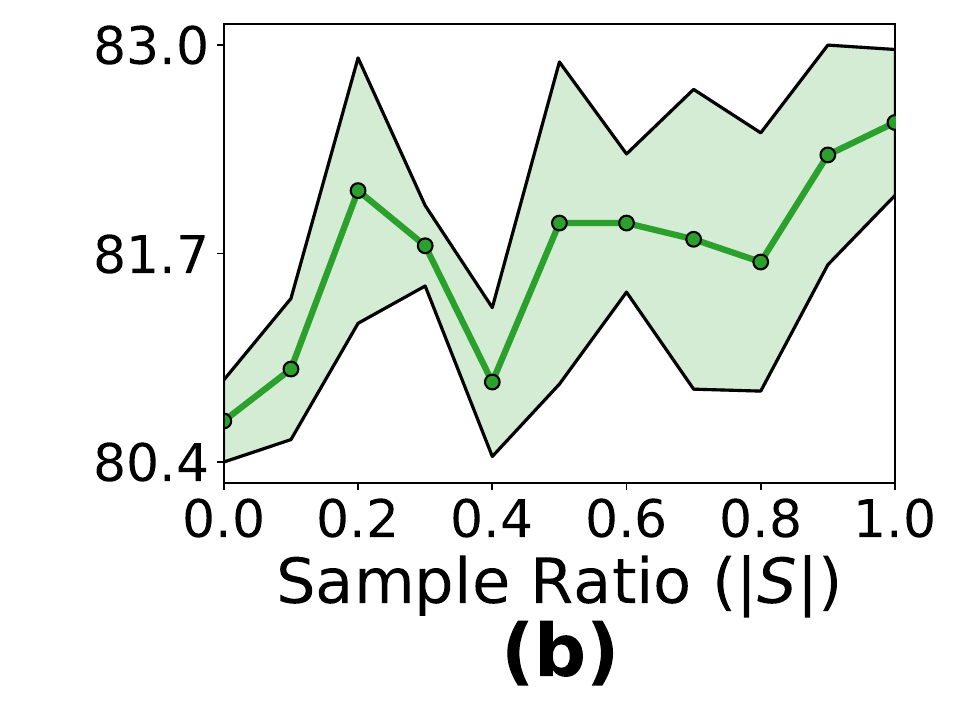}
    \end{minipage}
    \begin{minipage}[t]{0.28\linewidth}
        \centering        
        \includegraphics[width=\linewidth]{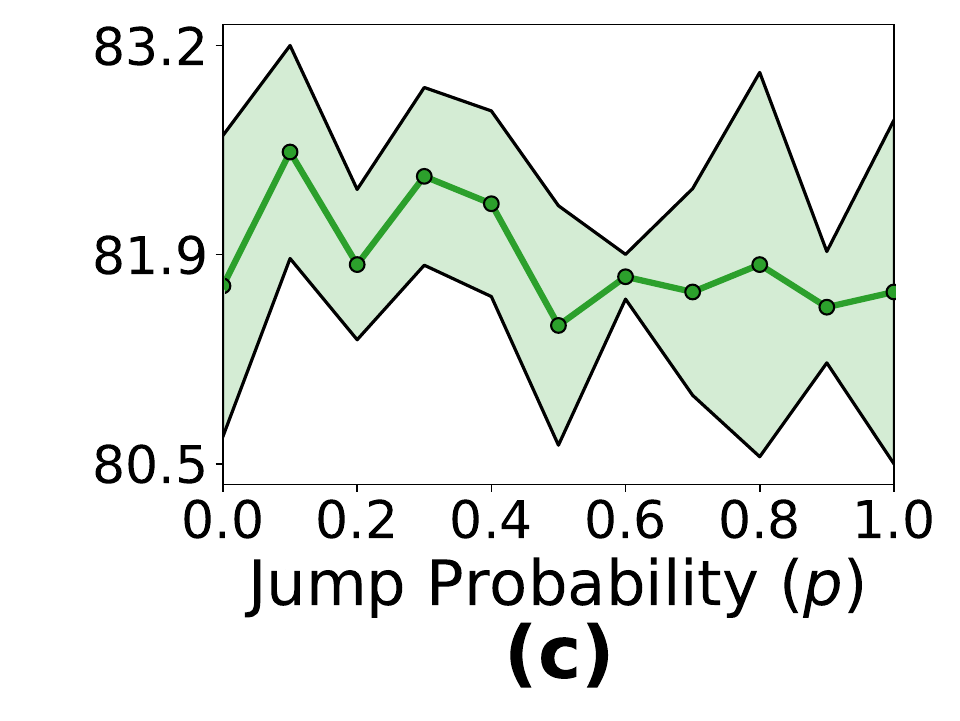}
    \end{minipage}

    \caption{Sensitivity of the walk against its hyperparameters with respect to the HoGA-GAT model, specifically: (a) the weighting factor $\gamma$ between history buffer and greedy step, (b) the percentage of nodes from $\lvert E_k \rvert$ sampled, and (c) the probability of jumping to a random node in $L_k(G)$. Shaded regions indicate one standard deviation from the mean.}
    \label{fig:bbbbb}
\end{figure*}
\begin{figure}[t]
    \centering

    \begin{minipage}[t]{0.49\linewidth}
        \centering
        \includegraphics[width=\linewidth]{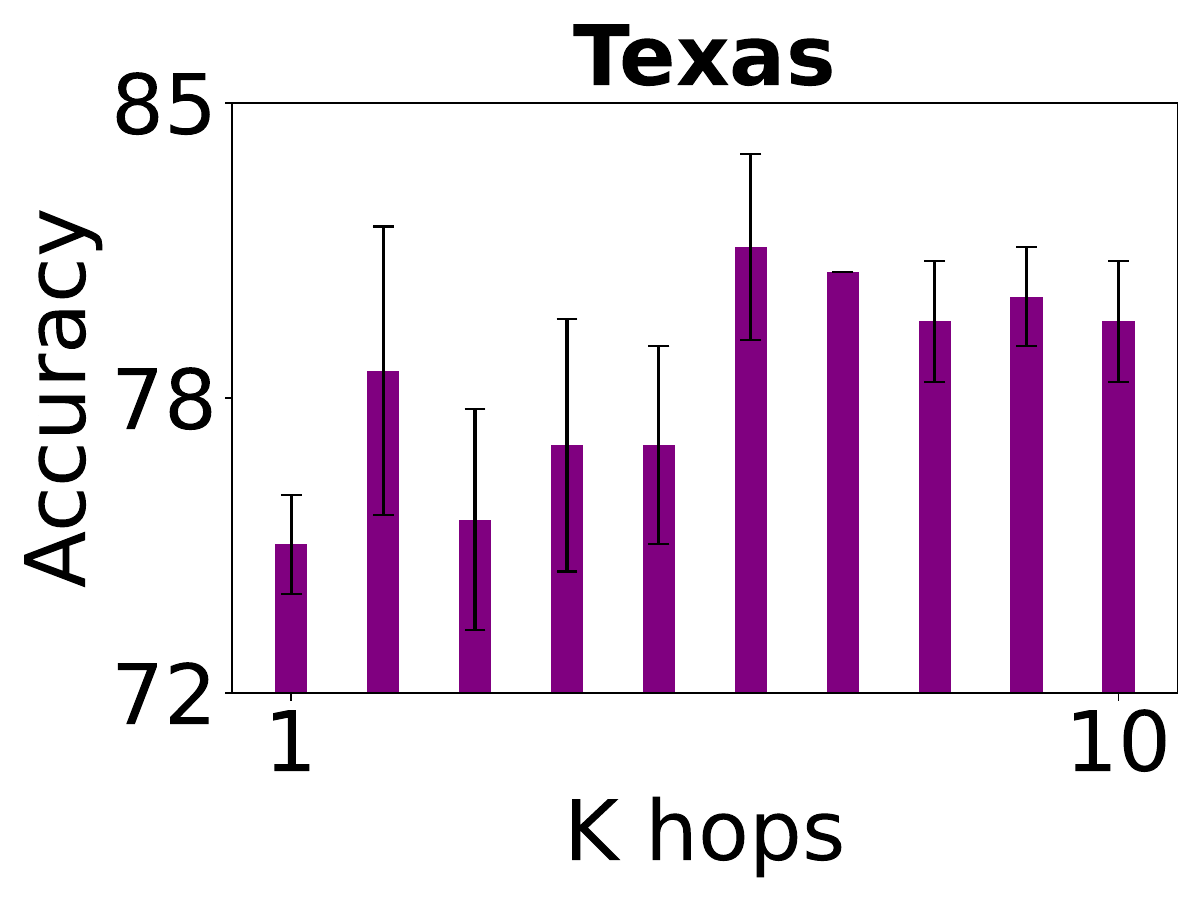}
    \end{minipage}\hfill
    \begin{minipage}[t]{0.455\linewidth}
        \centering
        \includegraphics[width=\linewidth]{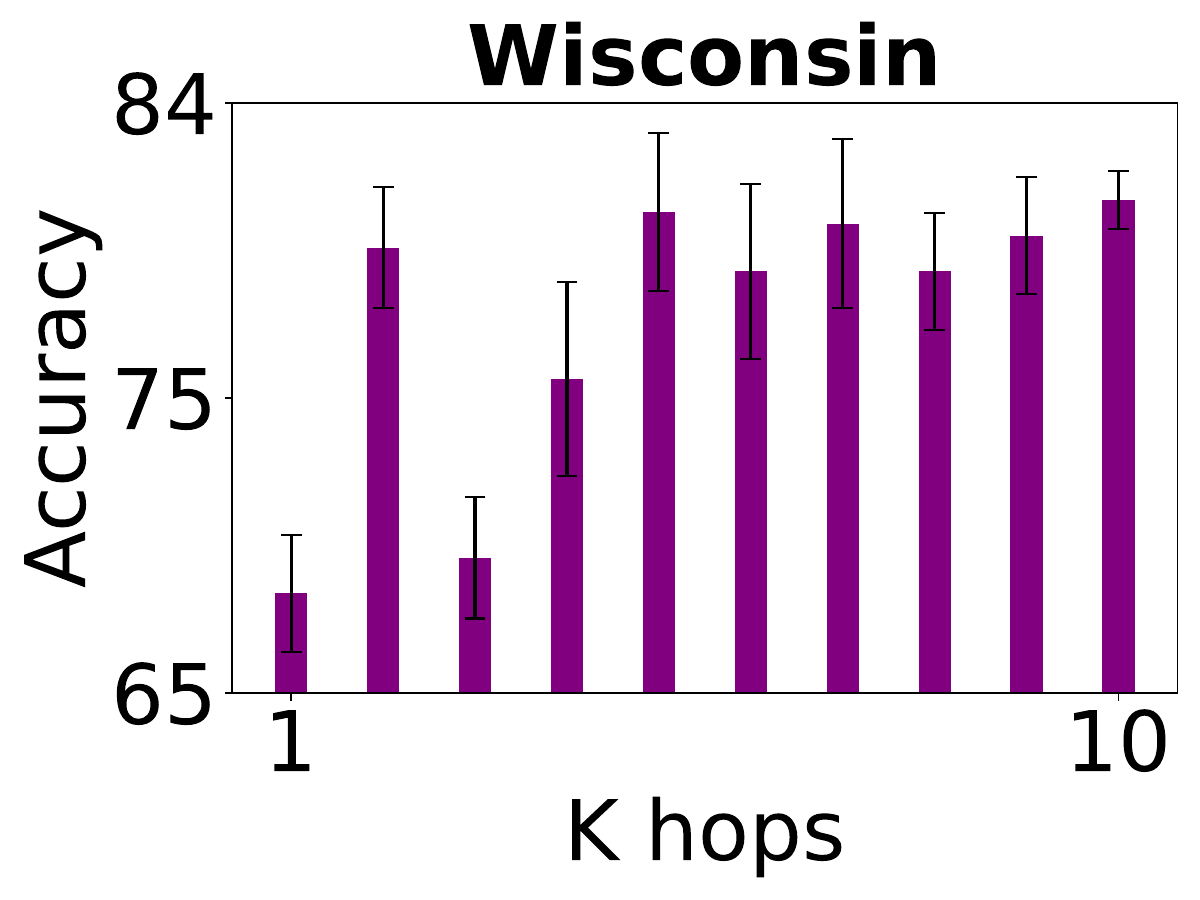}
    \end{minipage}

    \vspace{0.3em}

    \begin{minipage}[t]{0.49\linewidth}
        \centering
        \includegraphics[width=\linewidth]{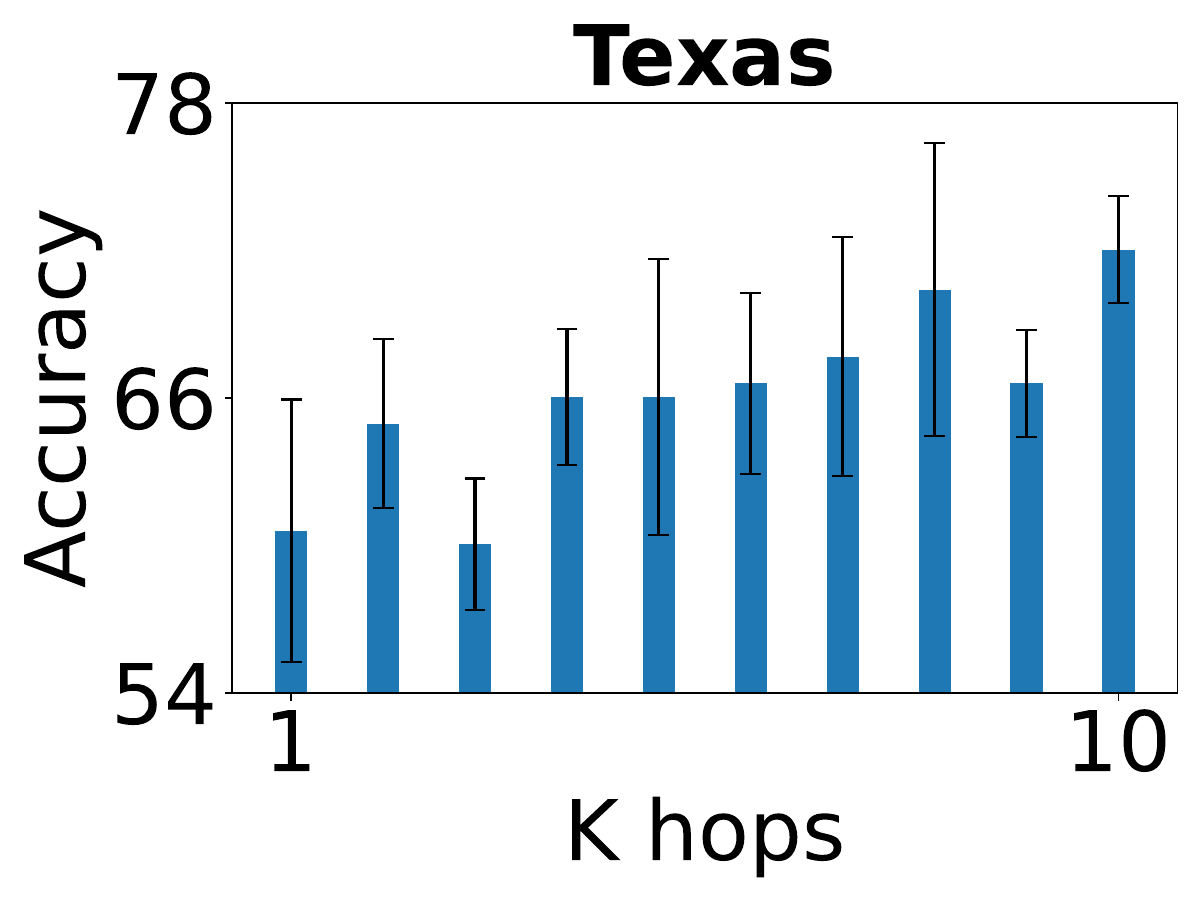}
    \end{minipage}\hfill
    \begin{minipage}[t]{0.46\linewidth}
        \centering
        \includegraphics[width=\linewidth]{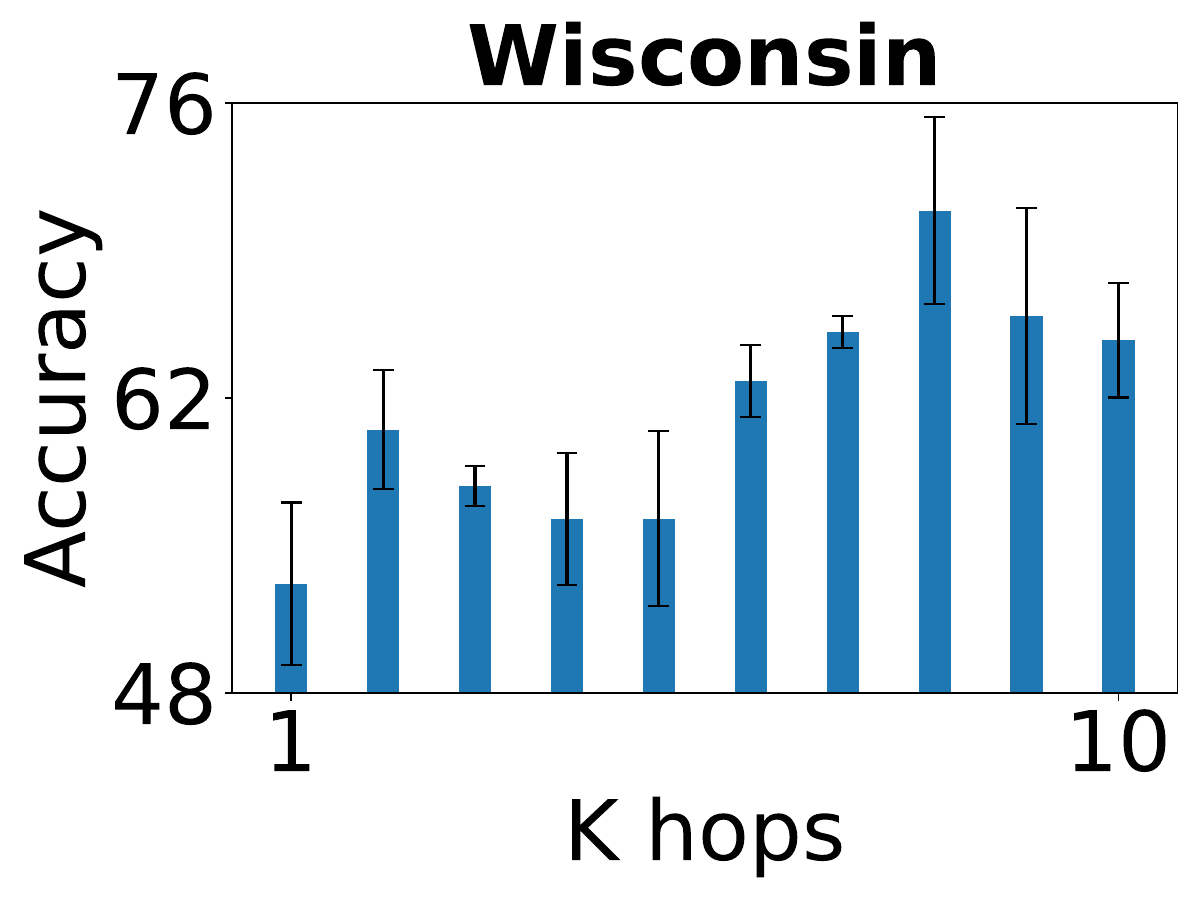}
    \end{minipage}

    \caption{Effects of varying the maximum hop for HoGA-GRAND (top) and HoGA-GAT (bottom) on two heterophilic datasets on which single-hop attention typically performs poorly.}
    \label{fig:gat_appendix_k}
\end{figure}

{\bf Testing Hop Number Stability (RQ2).} Figures \ref{fig:oversmooth_ab_test}(a) and \ref{fig:oversmooth_ab_test}(b) assess the stability of HoGA-GAT and HoGA-GRAND across different maximum hop numbers $K$, respectively. We observe that on Citeseer, Cora, and Pubmed, accuracy improves with small $K$ values, typically up to $ K=3$, after which the accuracy plateaus. Since these graphs have a diameter close to 3, additional hops mainly encode redundant information, which is the repetition of information already encoded at lower values of $k$. However, stability in $K$ indicates that our HoGA-GAT effectively retains higher-order information, allowing $K$ to be set to the graph diameter. The accuracy of HoGA-GRAND, however, typically decays after $3$-hops, indicating heightened sensitivity to the repeated information.

Figure \ref{fig:oversmooth_ab_test}(c) demonstrates the impact of scaling $\beta(k)$ on HoGA-GAT. On the Cora and PubMed, accuracy remains stable for scaling factors below 1.2; however, values overfit, and a subsequent drop in accuracy occurs. In contrast, Citeseer benefits more from incorporating long-distance information. Additionally, Figure \ref{fig:gat_appendix_k} shows the effect of maximal hop number on HoGA-GRAND and HoGA-GAT for graphs with a high heterophilic index, specifically Texas and Wisconsin. In all cases, performance over maximal hop number stably increases. Our findings suggest that HoGA counteracts the well-studied negative effects of heterophily \cite{wang2024understanding}, specifically by collating various diverse $k$-order relationships. 

{\bf Analysis on Oversmoothing (RQ2).} We use accuracy as a proxy metric, shown in Figure \ref{fig:oversmooth_ab_test}(d), to assess the degree of oversmoothing on node feature vectors caused by additional message-passing steps. Given that the diversity of feature vectors in the $k$-hop neighborhood increases with $k$ \cite{ai2024a2gcn}, increasing the $k$-order of an aggregation method enhances access to descriptive information. Intuitively, this mitigates the effect of positive-feedback structures, \textit{e.g.} homophilic cliques and cycles. Since HoGA-GAT aims to harness a subset of maximally diverse feature vectors from the $k$-hop neighborhoods, we observe in Figure \ref{fig:oversmooth_ab_test}(d) a reduction in the degree of oversmoothing for Citeseer and Cora. Despite the mitigation, performance consistently declines with additional message-passing steps. The decline may be due to vanishing gradients from increasing network depth \cite{hanin2018neural}. 

{\bf Sampling Methods Comparison (RQ3).} We evaluate the relative utility of sampling via feature vectors against search methods that capture graph topology strictly based on local connections, without optimizing for feature-vector diversity. We compare our heuristic walk with conventional, non-feature-vector-oriented methods: breadth and depth-first search, uniform random walk, and node selection. We also assess the importance of the history buffer by comparing it with a greedy walk. 

Table \ref{tab:node_sampling_table} shows our evaluation of these methods. We observe that the topology-oriented baselines acquire lower accuracy across all datasets, that is, a decrease of at least 2\%, 3\%, and 1\% on Cora, Citeseer, and Pubmed, respectively. Intuitively, the lack of inherent bias towards topological substructures, {\it e.g.} cliques organized via feature-vector similarity, leads to a less descriptive subset of the $k$-hop neighborhood; the walk-based search methods capture the entire localized substructure of the graph, causing the consequent subgraph topology to lack in descriptivity. Similarly, Random does not describe any causal relationships via edges and paths between nodes. Greedy, however, typically outperforms all methods except Heuristic Walk, which is due to its inability to consider previously seen global topology. Our heuristic walk more closely describes the global distribution of feature vectors.

\textbf{Comparing Performance with Walk Behavior (RQ3).} We extend the theoretical results of Theorem~\ref{th:th1} and Corollary~\ref{th:th2} through an empirical analysis of walk hyperparameters. In particular, Figure \ref{fig:bbbbb}\textbf{(a)} varies the weighting factor $\gamma$ between the history buffer and the greedy step. HoGA achieves peak performance at $\gamma = 0$, indicating the greedy step does not always offer benefit. Panel Figure \ref{fig:bbbbb}\textbf{(b)} shows that sampling more edges from the $k$-order line graph tends to increase accuracy, although this implies increased computational cost. In Figure \ref{fig:bbbbb}\textbf{(c)}, we vary the jump probability towards random nodes. Performance degrades as the walk resembles uniform sampling, except at low jump probabilities. The result indicates that limited random exploration helps the walk shift to new regions, improving coverage of the global graph structure. 

\section{Conclusion}

We proposed the Higher Order Graphical Attention (HoGA) module, which extends existing forms of single-hop self-attention methods to a $k$-hop setting. The simplicity of our method allows for both ease of implementation and applicability. In an empirical study, we show that HoGA significantly increases accuracy on node classification tasks across a range of benchmark datasets \cite{yang2016revisiting,shchur2018pitfalls} for both the GAT \cite{velickovic2018graph} and GRAND \cite{chamberlain2021grand} attention-based models. We also empirically demonstrate that direct sampling of the $k$-hop neighborhood is a strong competitor to other higher order methods \cite{wang2020multi,abboud2022shortest}, including topological \cite{huang2024higher,zhang2024hongat} and meta-path approaches \cite{yang2021heterogeneous,yang2021spagan}, while extending existing walk-based methods \cite{kong2022geodesic,michel2023path} to this setting.


\appendix

\section{Runtime and Memory Consumption}
Here, we analyze the asymptotic complexity of applying our HoGA model and present an empirical evaluation of its runtime and GPU memory usage.

\subsection{Discussion on Complexity}

We divide the complexity analysis into: (1) a review of the mathematical notation used in this paper, and (2) an examination of the growth of the parameter space and runtime complexity for the HoGA module. 

\textbf{Reminder on mathematical notation.} Recall that $G = (V, E)$ denotes a graph $G$ with vertex set $V$ and edge set $E$. Furthermore, that $G_k = (V, E_k)$ is the induced edge-altered graph of $G$, where for any edge $(i, j) \in E_k$ we have that there exists a path $\mathcal{P} = (i=i_1, \dots, j=i_k)$ of length $k$ between $i$ and $j$ consisting of vertices $i_1, \dots, i_k \in V$. 

\textbf{Parameterisation.} One motivation for our method is that each higher-order adjacency matrix $\mathbf{A}_k(\mathbf{x}(t), \mathcal{S}_k)$ contains $\mathcal{O}(\left\vert E \vert\right)$ non-zero entries. In the HoGA-GAT model variant, each edge is parameterized by a single-layer neural network, denoted as $a_{\theta_k}(\mathbf{x}_i(t), \mathbf{x}_j(t))$, for $(i, j) \in E_k$. Due to the limited sample size, parameter sharing is feasible. Thus, the total number of parameters introduced is $\sum_{1 \leq k \leq K} \text{dim}(\theta_k) = K \cdot \text{dim}(\theta_1)$, where $\text{dim}(\theta_k)$ is the dimensionality of the parameter-vector $\theta_k$. We assume homogeneity across all hop values; $\text{dim}(\theta_1)$ represents the original parameter vector from the single-hop attention module. Consequently, the number of parameters increases linearly with $\text{dim}(\theta_1)$ relative to $\theta_k$. The number of parameterized edges also increases at a similar asymptotic rate, which is $\mathcal{O}(K\cdot\left\vert E \vert\right)$.

\textbf{Runtime.} For the implementation of HoGA-GAT, we first perform matrix multiplication between $\mathbf{A}_k(\mathbf{x}(t), \mathcal{S}_k)$ and $\mathbf{x}(t)$. For HoGA-GRAND, we first calculate the complete adjacency matrix $\mathbf{A}_{1:K}(\mathbf{x}(t), \mathcal{S}_k)$. Given that $\mathbf{A}_k(\mathbf{x}(t), \mathcal{S}_k)$ is stored as an adjacency list, matrix multiplication is carried out in order $\mathcal{O}(\left\vert E \right\vert)$. 
In the computation of the attention matrix and the multiplication operation,
each additional hop requires step summarily requires order of $\mathcal{O}(2\cdot\left\vert E \right\vert)$ additional steps.

\textbf{Preprocessing}. We add an asymptotic complexity analysis of the Heuristic Walk to our revised manuscript. Let $G = (V, E)$ be a graph and $L_k(G) = (V, E_k)$ its $k$-order line graph. For each $k$, we precompute and store the shortest paths between $(i, j) \in E_k$ in $\mathcal{O}(|V| + |E_k|)$ time. Each feature comparison costs $\mathcal{O}(d)$ for feature dimension $d$, and each node has $\mathcal{O}(b^k)$ neighbors, where $b$ is the average degree. Thus, each sample requires $\mathcal{O}(d \cdot b^k)$ comparisons. With a sampling budget of $|E|$, the total complexity is $\mathcal{O}(d \cdot b^k \cdot |E| + |V| + |E_k|)$. HoGA scales to large graphs when $k$ is bounded by the graph diameter (typically 3-7). Note that this preprocessing step is performed only once. 



\section{Proofs of Theoretical Statements}

\setcounter{theorem}{0}

{\bf Proof of Theorem 1.} Suppose we have a walk $H_\tau$ consisting of $\tau$ nodes $i_{q_1}, \dots, i_{q_t}$ from the vertex set of a graph $G = (V, E)$. We write the probability of visiting every node in the cycle $C$, consisting of $L$ nodes $j_{u_1}, \dots, j_{u_L}$, in sequence as: 
    \begin{align}
        \mathbb{P}(i_{n+L} = j_{u_L}, \dots, i_n = j_{u_1} \mid C),
    \end{align}
    which uses the chain rule and is expressed as:
    \begin{align}
        \prod_{1 \leq l \leq L} \mathbb{P}(i_{n+l} = j_{u_l} \mid i_{n+l-1} = j_{u_{l-1}}, \dots, i_1 = j_{u_1}).
    \end{align}
    Next, we take Equation~\ref{eq:dissim} from the main paper as the normalized probability. We also express the contribution from the walk in terms of the constraint $\delta_q(H_\tau) \approx \frac{1}{r_{q,\tau}}$, where $r_{q,\tau}$ denotes the multiplicity of $q$ in $H_\tau$, to simplify the effects of the history buffer. The result is then followed by assigning edge weights $\omega_{i,j,\tau}$ to every $(i, j) \in E_k$ of $L_k(G)$, and updating per each iteration.

{\bf Proof of Corollary 1.} The result follows from supposing that for some $q \in C$, we also have $q \in H_\tau$ at some walk iteration with length $\tau$. The probability of leaving the cycle follows immediately from applying Theorem 1.  

\section{Explanation of Source Code}

This section discusses our source code, how to run the models, and how to conduct experiments such as an ablation test. The results of our hyperparameter searches and other experiments are in the source code. 

\noindent\textbf{Generic running signature.} To run a model, use: 

\texttt{python main.py} \\
\texttt{\phantom{python }--dataset dataset\_name} \\
\texttt{\phantom{python }--train} \\
\texttt{\phantom{python }--model model\_name}

\noindent where \texttt{model\_name} and \texttt{dataset\_name} are selections from the supported models and datasets. On completion, the model, the average metrics over a specified number of runs, and a file containing the model's hyperparameters will be saved to memory. You can retrieve the results by replacing the \texttt{--train} flag with \texttt{--test}. In any case, metric results from each model seed will also be displayed via terminal output.

\begin{acks}
This research is supported by MBIE Strategic Science Investment Fund (SSIF) Data Science platform - Time-Evolving Data Science / Artificial Intelligence for Advanced Open Environmental Science (UOWX1910).
\end{acks}

\bibliographystyle{ACM-Reference-Format}
\bibliography{wsdm}

\end{document}